\documentclass[runningheads]{llncs}

\usepackage{eccv}

\usepackage{eccvabbrv}

\usepackage{graphicx}
\usepackage{multirow}
\usepackage{booktabs}
\usepackage{rotating}
\usepackage{siunitx} 
\usepackage{makecell}

\def\eg{\emph{e.g}\onedot} 
\def\ie{\emph{i.e}\onedot}

\usepackage[accsupp]{axessibility}  

\usepackage{hyperref}
\usepackage{orcidlink}

\begin{document}
\newcommand{\modelname}{Render-FM}
\title{\modelname: Feedforward Model for Real-time Photorealistic Volumetric Rendering} 

\titlerunning{Render-FM: Feedforward Model for Volumetric Rendering}

\author{Zhongpai Gao\thanks{Corresponding author}\orcidlink{0000-0003-4344-4501} \and
Benjamin Planche\orcidlink{0000-0002-6110-6437} \and
Meng Zheng\orcidlink{0000-0002-6677-2017} \and
Anwesa Choudhuri\orcidlink{0009-0005-7342-0182} \and
Van Nguyen Nguyen\orcidlink{0000-0003-4749-9040} \and
Terrence Chen\orcidlink{0009-0001-6697-7098} \and
Ziyan Wu\orcidlink{0000-0002-9774-7770}}

\authorrunning{Z.~Gao et al.}

\institute{United Imaging Intelligence, Boston, MA, USA\\
\email{\{zhongpai.gao, benjamin.planche, meng.zheng, anwesa.choudhuri, vannguyen.nguyen, terrence.chen, ziyan.wu\}@uii-ai.com}}

\maketitle

\begin{figure}[h]
\begin{center}
\includegraphics[width=1.\textwidth]{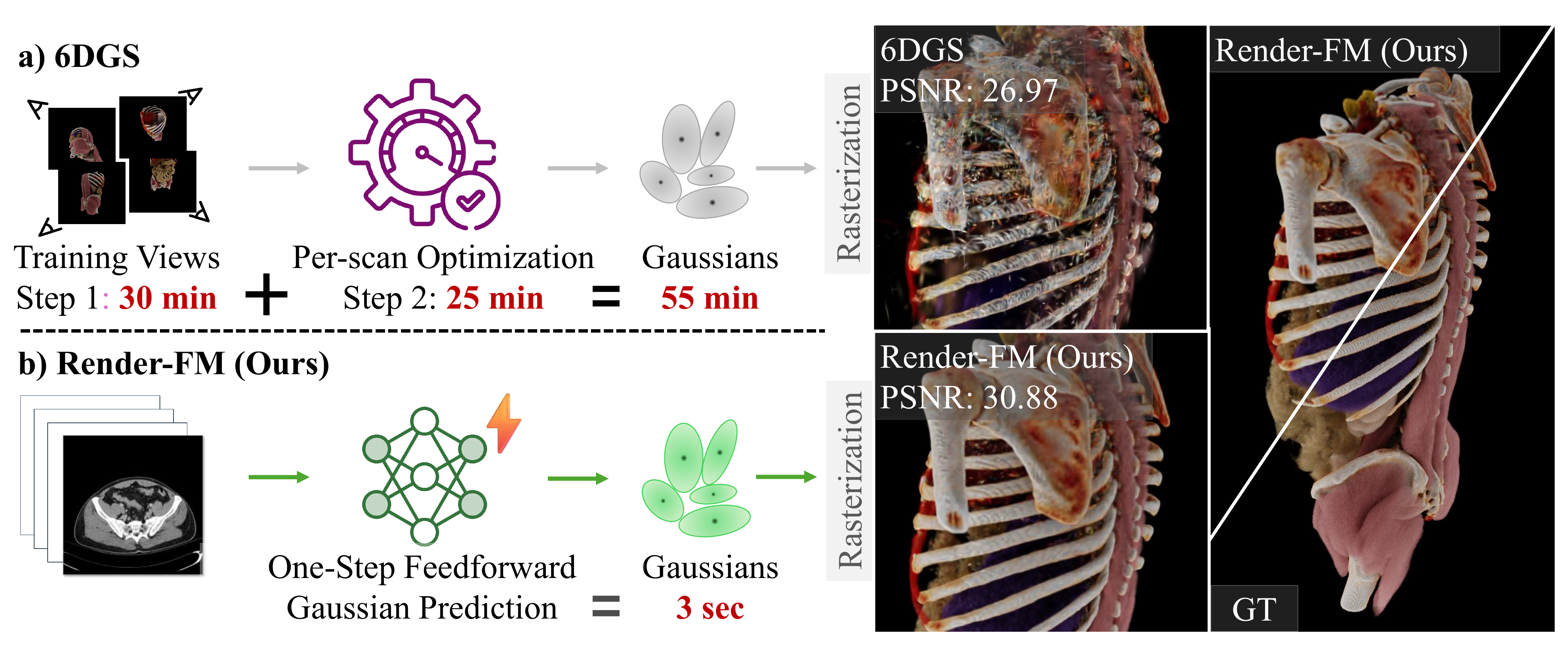}
\end{center}
\vspace{-2em}
\caption{Comparison of CT scan volumetric rendering pipelines. \textbf{Top:} 6DGS \cite{6DGS_arxiv_39} relies on physically-based rendering (PBR) of training views followed by per-scan optimization, and exhibits severe artifacts under limited training views. \textbf{Bottom:} Our \modelname{} produces high-quality renderings via a single feedforward pass.}
\vspace{-3em}
\end{figure}

\begin{abstract}
  Photorealistic volumetric rendering of CT scans greatly benefits clinical workflows, yet neural approaches such as Neural Radiance Fields (NeRF) and 3D Gaussian Splatting (3DGS) require prohibitive per-scan optimization (hours for NeRF, about 30 minutes for 3DGS), making them impractical in clinical settings. We propose \modelname{}, a feedforward model that eliminates this bottleneck by directly regressing 6D Gaussian Splatting (6DGS) parameters from a CT volume in a single 2.8-second forward pass, a 500$\times$ speedup over per-scan optimization. To bridge the domain gap between natural scene reconstruction and medical volumetric rendering, we introduce Anatomy-Guided Priming (AGP), which incorporates segmentation masks and transfer functions as structural and appearance priors, information that existing Gaussian splatting methods overlook. Built on an nnU-Net-inspired 3D U-Net trained on diverse CT scans, \modelname{} predicts per-voxel 6DGS parameters and supports immediate real-time rendering. Unlike per-scan methods, it generalizes to unseen anatomies, novel transfer functions, and enables compositional organ visualization with zero additional preparation time. Optional 89-second fine-tuning further improves quality, surpassing per-scan optimized baselines. Project page: \url{https://gaozhongpai.github.io/renderfm/}.
\end{abstract}

\section{Introduction}
\label{sec:introduction}

Medical imaging modalities like Computed Tomography (CT) produce rich volumetric datasets that are conventionally reviewed as sequences of 2D slices. While fundamental to clinical practice, this slice-by-slice approach often fails to convey the intricate spatial relationships of 3D anatomical structures and pathologies, particularly in complex cases involving multi-organ interactions or vascular networks \cite{dappa2016cinematic}. Volumetric rendering addresses this limitation by synthesizing comprehensive 3D views that enable intuitive and interactive exploration, significantly enhancing diagnostic assessment, surgical planning, and patient communication \cite{beyer2007high, eid2017cinematic}. The ability to dynamically inspect patient anatomy from arbitrary perspectives fundamentally transforms how clinicians interact with medical imaging data \cite{caton2020three}.

Despite technological advancements, achieving photorealistic volumetric rendering faces significant barriers to routine clinical adoption. While conventional Direct Volume Rendering (DVR) methods \cite{jung2019direct} enable real-time interaction through GPU-accelerated ray-casting, they employ simplified local illumination models that produce visually limited results. In contrast, photorealistic rendering techniques like Cinematic Rendering \cite{dappa2016cinematic, eid2017cinematic, wollschlaeger2020ct, siemens_cinematic_rendering} and physically-based path tracing achieve remarkable visual quality through global illumination, soft shadows, and subsurface scattering, effects critical for realistic tissue appearance and enhanced spatial perception in surgical planning and patient communication. However, these methods require computationally expensive light transport simulation. More recent approaches, such as Neural Radiance Fields (NeRF) \cite{mildenhall2021nerf} and 3D Gaussian Splatting (3DGS) \cite{kerbl20233dgs}, can achieve photorealistic quality but impose prohibitive per-scan optimization requirements that fundamentally conflict with clinical time constraints.

The core bottleneck lies in per-scene optimization: 3DGS typically requires around 30 minutes, while NeRF can extend to 10+ hours. This is compounded by the prerequisite generation of physically-rendered training views, consuming approximately 18 seconds per view, bringing the full preparation pipeline to nearly an hour per case for 3DGS, and far longer for NeRF. Beyond speed, optimized parameters cannot transfer between scans due to variations in anatomy, pathology, and acquisition protocols, nor can they generalize to new transfer functions (\ie, color and opacity definitions) or support compositional organ visualization, capabilities essential for comprehensive diagnostic workflows.

Recent advances in large-scale feedforward models present a transformative paradigm for addressing these limitations. Inspired by foundation model pre-training principles \cite{FoundationModelsGeneral, moor2023foundation, paschali2025foundation}, Large Gaussian Models (LGMs) \cite{tang2024lgm, xu2024grm} have demonstrated that feedforward networks trained on large datasets can directly predict 3D Gaussian parameters from sparse image inputs, bypassing per-scene optimization entirely. While LGMs target natural scene reconstruction from 2D views, their core insight transfers naturally to medical volumetric rendering: CT scans provide dense, structured 3D input that makes the feedforward mapping more tractable, and large-scale CT datasets spanning diverse institutions and pathologies offer the breadth needed to learn generalizable anatomical priors across patients.

We introduce \modelname{}, a feedforward model for photorealistic volumetric rendering through direct prediction of 6D Gaussian Splatting (6DGS) parameters from CT volumes. Unlike optimization-based methods requiring extensive per-scan training, \modelname{} performs parameter regression in a single 2.8-second forward pass, a 500$\times$ speedup, while maintaining superior visual quality. We adopt 6DGS \cite{6DGS_arxiv_39} over standard 3DGS for its explicit modeling of complex view-dependent optical effects in a 6D spatio-angular space, better capturing scattering and reflectance phenomena at tissue interfaces that are critical for photorealistic medical visualization. The model employs an encoder-decoder architecture inspired by nnU-Net's medical imaging principles \cite{nnUNet_NatureMethods_43, isensee2024nnu}, combined with our novel Anatomy-Guided Priming (AGP) that incorporates anatomical segmentation masks and transfer functions as structured priors, contextual information that existing Gaussian Splatting methods entirely overlook.

Trained on diverse CT datasets spanning multiple institutions and pathologies \cite{wasserthal2023totalsegmentator}, \modelname{} learns robust generalization across clinical imaging conditions. The resulting model enables immediate real-time rendering at 328+ FPS while supporting dynamic transfer function modification and compositional organ visualization, capabilities impossible with optimization-based approaches. Optional fine-tuning can further enhance quality to 31.67 dB PSNR within 89 seconds, providing flexibility while maintaining practical deployment timelines. Our key contributions include:
\begin{itemize}
    \item \textbf{Feedforward Model Architecture:} A novel feedforward model integrating nnU-Net's medical imaging principles with 6DGS's view-dependent rendering capabilities to directly regress 6DGS parameters from CT volumes, eliminating per-scan optimization and reducing preparation time from ${\sim}$1 hour to 2.8 seconds.
    \item \textbf{Anatomy-Guided Priming (AGP):} A novel initialization strategy that leverages segmentation masks and transfer functions to provide anatomically-informed structural and appearance priors for 6D Gaussian primitives, bridging the domain gap between natural scene reconstruction and medical volumetric rendering.
    \item \textbf{End-to-End Training \& Evaluation:} A comprehensive training methodology using differentiable 6DGS rendering on large-scale CT datasets, with extensive evaluation demonstrating superior rendering quality, 500$\times$ speedup over per-scan optimization, and robust generalizability to unseen anatomies, novel transfer functions, and compositional organ visualization.
\end{itemize}

\section{Related Work}
\label{sec:related_work}

\noindent\textbf{Volumetric Rendering in Medical Imaging} $\,$ Direct Volume Rendering (DVR) synthesizes images by casting rays through volumetric data using transfer functions to map voxel intensities to optical properties \cite{heng2006gpu, jung2019direct}. Standard GPU-accelerated DVR implementations enable real-time interaction in clinical viewers like 3D Slicer and OsiriX \cite{beyer2007high}, but employ simplified local illumination (\eg, Phong shading) producing visually limited results. Cinematic Rendering \cite{dappa2016cinematic, eid2017cinematic, elshafei2019comparison, siemens_cinematic_rendering} achieves photorealism through physically-based rendering with global illumination, soft shadows, and subsurface scattering. However, such photorealistic methods require expensive light transport simulation (seconds per view) and expert transfer function tuning, limiting routine clinical adoption despite their superior visual quality.

\noindent\textbf{Neural Radiance Fields (NeRF) and 3D Gaussian Splatting (3DGS)} $\,$ NeRF \cite{mildenhall2021nerf} implicitly represents scenes using MLPs that map 5D position-and-direction coordinates to density and color, achieving high-quality novel view synthesis at the cost of extensive per-scene optimization (hours to days) and slow rendering. 3DGS \cite{kerbl20233dgs} addresses rendering speed by explicitly modeling scenes with 3D Gaussian primitives and a differentiable tile-based rasterizer, enabling real-time rendering, but still requires around 30 minutes of per-scene optimization. Both have been adapted for medical imaging, including sparse-view CT reconstruction \cite{NeRF_Medical_Sparse_34} and X-ray visualization \cite{gao2024ddgs}. However, these adaptations retain the per-scan optimization requirement, and optimized parameters cannot transfer across scans or generalize to new transfer functions, which are fundamental limitations for clinical deployment.

\noindent\textbf{6D Gaussian Splatting} $\,$
6D Gaussian Splatting (6DGS) \cite{6DGS_arxiv_39} extends 3DGS by representing primitives within a 6D spatio-angular space, characterized by a 6D covariance matrix that captures variance in both 3D position and 3D direction. This formulation enables explicit modeling of complex view-dependent effects, such as anisotropic reflections and subsurface scattering, that are frequently observed at tissue interfaces in medical data and are difficult to capture with standard 3DGS spherical harmonics. During rendering, the 6D covariance is dynamically sliced based on the viewing direction to yield an effective 3D Gaussian for rasterization, achieving higher fidelity with potentially fewer primitives \cite{6DGS_arxiv_39}. These properties make 6DGS particularly well-suited for photorealistic medical visualization.

\noindent\textbf{Feedforward Models and Large Gaussian Models} $\,$ Foundation models pre-trained on large-scale datasets demonstrate robust generalization to downstream tasks without task-specific optimization \cite{FoundationModelsGeneral, moor2023foundation, paschali2025foundation}. Large Gaussian Models (LGMs) \cite{tang2024lgm, xu2024grm} bring this paradigm to 3D reconstruction, training feedforward networks to directly predict 3D Gaussian parameters from sparse 2D image inputs without per-scene optimization. However, LGMs are designed for natural scene reconstruction from RGB images, a setting fundamentally different from medical volumetric rendering, which involves structured 3D CT input, domain-specific appearance priors (transfer functions), and the need for anatomical generalization across patients. \modelname{} adapts the feedforward paradigm to this medical setting by leveraging dense 3D CT volumes as input and incorporating anatomy-guided priors, addressing the gaps that make LGMs ill-suited for direct clinical application.

\noindent\textbf{nnU-Net Framework} $\,$
The nnU-Net framework \cite{nnUNet_NatureMethods_43, isensee2024nnu} excels in medical image segmentation through self-configuring pipeline adaptation, automatically adjusting patch size, normalization, and architecture to handle the wide variation in CT resolutions, spacings, and fields-of-view across institutions. Its 3D U-Net backbone consistently achieves state-of-the-art results across diverse benchmarks, and its principles have been extended beyond segmentation to tasks such as landmark detection \cite{nnLandmark_arxiv_41} and interactive segmentation \cite{isensee2025nninteractive}. \modelname{} leverages these same architectural principles for CT-to-6DGS parameter regression, benefiting from nnU-Net's robustness to clinical CT variability.

To our knowledge, \modelname{} is the first method to combine feedforward prediction, 6D Gaussian Splatting, and anatomy-guided priming for photorealistic CT volumetric rendering, eliminating per-scan optimization while achieving rendering quality comparable to or better than per-scan optimized baselines.

\section{Methodology}
\label{sec:methodology}

\begin{figure}[t]
\centering
\includegraphics[width=\linewidth]{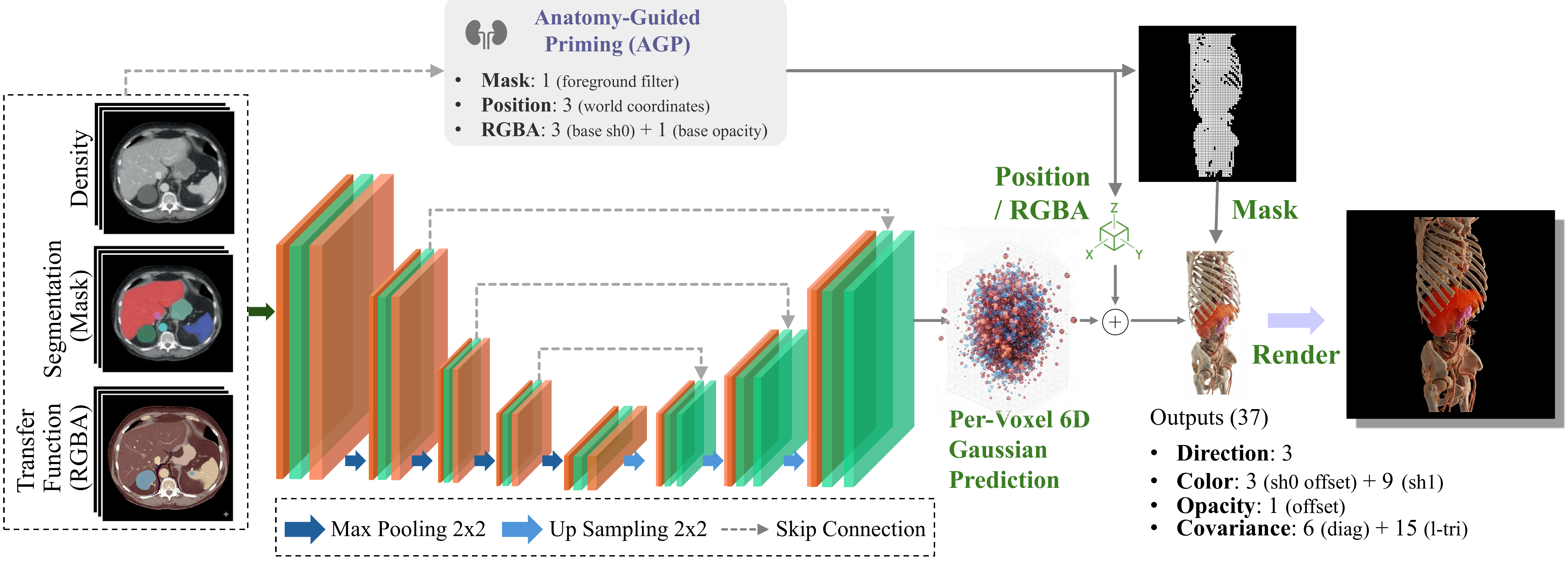}
\caption{Overview of the \modelname{} pipeline. A 3D U-Net encoder-decoder network processes a 6-channel input volume and regresses 37-channel 6DGS parameters per voxel. Foreground voxels instantiate 6D Gaussians, which are rendered via differentiable view-dependent slicing and tile-based rasterization. End-to-end training is supervised by a rendering loss against physically-based ground-truth views.}
\label{fig:architecture}
\end{figure}

\modelname{} learns a direct mapping from a CT volume to 6D Gaussian Splatting (6DGS) parameters via a single feedforward pass, enabling real-time photorealistic rendering without per-scan optimization. An overview of the pipeline is shown in Fig.~\ref{fig:architecture}.

\paragraph{\textbf{Problem Formulation}} Given a 3D CT scan $V \in \mathbb{R}^{C \times D \times H \times W}$ with $C=6$ input channels, our goal is to predict parameters $\Theta$ defining a 6DGS representation that enables high-quality, view-dependent rendering from arbitrary viewpoints. Formally, we learn a function $f_\phi: V \mapsto \Psi$, where $\Psi \in \mathbb{R}^{37 \times \lfloor D/2 \rfloor \times \lfloor H/2 \rfloor \times \lfloor W/2 \rfloor}$ represents voxel-wise 6DGS parameters across the volume, parameterized by network weights $\phi$. During rendering, a foreground subset $\Theta \subset \Psi$ determined by the segmentation mask is instantiated as 6D Gaussian primitives.

\subsection{Model Architecture}

\noindent\textbf{Input Representation} $\,$
The 6-channel input $V$ provides complementary anatomical and appearance information: 1) normalized CT intensity (Hounsfield Units), capturing tissue density; 2) a segmentation mask identifying foreground structures; and 3) four RGBA channels from pre-defined transfer functions, encoding base color and opacity per anatomy class. This multi-channel design enriches the network with structural and appearance context, including tissue density, semantic identity, and base appearance, reducing the burden of learning these priors from scratch and enabling more accurate Gaussian parameter prediction.

\vspace{0.4em}
\noindent\textbf{Encoder-Decoder Backbone} $\,$
\modelname{} employs a 3D U-Net architecture inspired by nnU-Net \cite{nnUNet_NatureMethods_43, isensee2024nnu}. The encoder applies successive blocks of two $3\times3\times3$ convolutions with Instance Normalization and Leaky ReLU activation, interleaved with $2\times2\times2$ max-pooling for downsampling. The decoder mirrors this structure with transposed convolutions for $2\times2\times2$ upsampling and skip connections that concatenate encoder features at corresponding resolutions (excluding the final stage, which outputs at half the input resolution). The final $1\times1\times1$ convolution produces 37 output channels per voxel, yielding $\Psi \in \mathbb{R}^{37 \times \lfloor D/2 \rfloor \times \lfloor H/2 \rfloor \times \lfloor W/2 \rfloor}$. To satisfy the spatial divisibility requirements of the multi-scale pooling operations, input volumes are zero-padded to the nearest multiple of 32 along each spatial dimension prior to the forward pass.

\vspace{0.4em}
\noindent\textbf{6DGS Parameter Prediction} $\,$
Each spatial location in $\Psi$ encodes the attributes of one potential 6D Gaussian primitive. The 3D position $\mu_p$ is fixed to the voxel's world-space coordinates and is not predicted by the network. The 37 predicted channels at each location encode: mean direction $\mu_d$ (3 channels), view-dependent color via Spherical Harmonic coefficients at degrees $L{=}0$ and $L{=}1$ (12 channels total), an opacity offset (1 channel), and the 21 upper-triangular elements of the $6\times6$ covariance matrix $\Sigma$. To guarantee a valid positive semi-definite covariance, the 6 diagonal elements are predicted in log-space, while the 15 off-diagonal elements are constrained via $\tanh$ activation before Cholesky reconstruction.

\subsection{Anatomy-Guided Priming}

Standard 3DGS initialization relies on SfM-derived sparse point clouds \cite{schoenberger2016sfm} or random placement, both unsuitable for volumetric medical data as they fail to leverage domain-specific anatomical information. Recent CT-based methods such as DDGS-CT \cite{gao2024ddgs} and 6DGS \cite{6DGS_arxiv_39} improve upon this by using marching-cubes on CT radiodensity, but still ignore segmentation labels, transfer function colors, and opacity, information that is readily available in clinical pipelines. We introduce Anatomy-Guided Priming (AGP) as a comprehensive initialization strategy for 6D Gaussian primitives. Since \modelname{} predicts a 6D Gaussian for each voxel, Gaussian positions are set directly to voxel world coordinates; base colors and opacities are derived from pre-defined transfer functions; and semantic labels are assigned from the segmentation mask. This anatomically-informed starting point guides the network to predict residual refinements on top of meaningful priors, rather than learning geometry and appearance entirely from scratch, leading to more stable training and better generalization.

\subsection{Differentiable Rendering}

\noindent\textbf{Gaussian Instantiation} $\,$
Since the network outputs $\Psi$ at half the input resolution, each foreground voxel index in $\Psi$ is first upsampled by a factor of 2 to recover full-resolution voxel coordinates, which are then mapped to physical world-space positions via the volume's affine transformation. Direction $\mu_{d,i}$ and covariance components are extracted from $\Psi$ at the corresponding half-resolution location, and the $6\times6$ covariance matrix $\Sigma_i$ is reconstructed from its predicted Cholesky factors. The appearance attributes are: color $c_i$, computed by combining AGP base RGB values with predicted spherical harmonic offsets; opacity $\alpha_i = \sigma(\alpha_{\text{base}} + \Delta\alpha_i)$, which blends the AGP prior with the predicted residual to ensure $\alpha_i \in [0,1]$; and class label $l_i$, assigned from the segmentation mask.

\vspace{0.4em}
\noindent\textbf{Covariance Slicing for View-Dependence} $\,$
For rendering from a camera at position $\mathbf{p}$, we apply view-dependent covariance slicing to each Gaussian $i \in \Theta$ as described in \cite{6DGS_arxiv_39}. The view direction is $\mathbf{v}_i = (\mu_{p,i} - \mathbf{p}) / \|\mu_{p,i} - \mathbf{p}\|$. The $6\times6$ covariance $\Sigma_i$ is partitioned into spatial ($pp$), directional ($dd$), and cross-term ($pd$, $dp$) blocks:
\begin{equation}
\Sigma_i = \begin{pmatrix} \Sigma_{pp} & \Sigma_{pd} \\ \Sigma_{dp} & \Sigma_{dd} \end{pmatrix}.
\end{equation}
The view-conditioned spatial mean and covariance are then:
\begin{align}
\mu'_{p,i} &= \mu_{p,i} + \Sigma_{pd} \Sigma_{dd}^{-1} (\mathbf{v}_i - \mu_{d,i}), \\
\Sigma'_{pp,i} &= \Sigma_{pp} - \Sigma_{pd} \Sigma_{dd}^{-1} \Sigma_{dp},
\end{align}
with an opacity modulation factor $w_i = \mathcal{N}(\mathbf{v}_i \mid \mu_{d,i}, \Sigma_{dd})$ that scales visibility based on the alignment between the viewing direction and the Gaussian's directional distribution. This slicing mechanism adapts each Gaussian's shape, position, and opacity per viewpoint, capturing view-dependent optical effects such as subsurface scattering and specular reflectance at tissue interfaces.

\vspace{0.4em}
\noindent\textbf{Tile-Based Rasterization} $\,$
The view-conditioned Gaussians (defined by $\mu'_{p,i}$, $\Sigma'_{pp,i}$, modulated opacity $w_i\alpha_i$, and color from $c_i$) are rendered using a differentiable tile-based rasterizer adapted from 3DGS \cite{kerbl20233dgs}. Gaussians are projected onto the image plane, sorted by depth, culled, and assigned to screen tiles. Alpha compositing is performed efficiently within each tile in parallel on the GPU. This fully differentiable process enables end-to-end training through the renderer and achieves real-time rendering speeds at inference.

\subsection{Training}

A central design choice of \modelname{} is the unification of the nnU-Net \cite{nnUNet_NatureMethods_43, isensee2024nnu} volumetric prediction backbone with the differentiable 6DGS rendering pipeline into a single end-to-end trainable system. Rather than treating parameter regression and rendering as separate stages, rendering gradients flow directly back through the rasterizer into the network weights, allowing the network to learn 6DGS parameters that are optimal for visual quality rather than for any intermediate proxy loss.

Concretely, at each training step, predicted Gaussians are rendered from 4 randomly sampled viewpoints $\mathbf{p}$ via differentiable rasterization $\hat{I} = R(\Theta, \mathbf{p})$ and compared against physically-based ground-truth images $I_{gt}$ produced offline via PBRT \cite{pharr2023physically}. We supervise training with a combination of pixel-level and perceptual losses:
\begin{equation}
\mathcal{L} = \lambda_{L1} \mathcal{L}_{L1} + \lambda_{\text{SSIM}} \mathcal{L}_{\text{SSIM}},
\end{equation}
where $\mathcal{L}_{L1} = \|\hat{I} - I_{gt}\|_1$ penalizes pixel-level error and $\mathcal{L}_{\text{SSIM}} = 1 - \text{MS-SSIM}(\hat{I}, I_{gt})$ \cite{SSIM_Metric} enforces structural fidelity, with weights $\lambda_{L1} = 0.8$ and $\lambda_{\text{SSIM}} = 0.2$. View losses are averaged before backpropagation across all training scans.

\subsection{Inference Pipeline}

The \modelname{} inference pipeline first preprocesses the CT volume: it is resampled to isotropic \SI{1.5}{\milli\metre} spacing, a segmentation mask is either provided or generated automatically (\eg, via TotalSegmentator \cite{wasserthal2023totalsegmentator}), and per-voxel RGBA values are computed from pre-defined transfer functions. The normalized CT intensity, segmentation mask, and RGBA volume are assembled into the 6-channel input $V$, zero-padded to the nearest multiple of 32.

The assembled input is passed through $f_\phi$ in a single forward pass, producing the dense parameter volume $\Psi$. Foreground voxels identified by the segmentation mask are immediately instantiated as 6D Gaussians with geometry and appearance attributes derived from $\Psi$ and the AGP priors, as described above. These 6D Gaussians are ready for real-time interactive rendering without any further optimization, enabling the clinician to inspect the anatomy from arbitrary viewpoints at 328+ FPS. The entire process from raw CT to interactive visualization completes in under 3 seconds on GPU, compared to approximately one hour required by per-scan optimization methods.

For cases requiring higher rendering fidelity, the instantiated 6DGS model can optionally be fine-tuned for less than 2 minutes. This fine-tuning optimizes the predicted Gaussian parameters directly on the test scan using the same differentiable renderer and loss, providing a quality boost while remaining far more practical than full per-scan optimization. The two-stage design, fast feedforward prediction followed by optional lightweight refinement, offers a flexible trade-off between speed and quality to suit different clinical demands.

\section{Experiments}
\label{sec:experiments}

We conduct extensive experiments to evaluate the performance of \modelname{} against the 6DGS baseline \cite{6DGS_arxiv_39}, focusing on rendering quality, computational efficiency, and generalizability. We first ablate the role of anatomy-guided priming (AGP) as both an initialization strategy and a necessary condition for feedforward learning, then present a full quantitative and qualitative comparison across in-domain and out-of-domain settings, including seen and unseen transfer functions, fine-tuning, and compositional organ visualization.

\subsection{Experimental Protocol}

\noindent\textbf{Datasets} $\,$
We utilized two publicly available CT datasets: 1) \textbf{TotalSegmentator \cite{wasserthal2023totalsegmentator}:} This dataset comprises 1,228 CT scans from clinical routines, covering 117 anatomical classes. It includes a wide range of pathologies, scanners, acquisition protocols, and institutions, making it representative of real-world clinical variability. After filtering scans exceeding 48 million voxels (due to GPU memory constraints) or lacking orthonormal directionality, we used 991 scans for training and 46 for in-domain (\textit{ID}) testing; 2) \textbf{CT-ORG \cite{rister2020ct}:} This dataset includes 140 CT scans from diverse sources, featuring both large organs (\eg, lungs) and small, challenging structures (\eg, bladder). We selected 10 scans (volumes 2--11) for out-of-domain (\textit{OOD}) testing to evaluate \modelname{}'s generalizability to unseen data distributions.

\vspace{0.4em}
\noindent\textbf{Data Preparation} $\,$
To prepare the training data, we applied a standardized preprocessing pipeline: 1) \textit{Resampling:} Volumes were resampled to isotropic spacing of \SI{1.5}{\milli\metre} in all dimensions to ensure consistency; 2) \textit{Normalization:} CT intensities (Hounsfield Units) were normalized following the nnU-Net pipeline \cite{nnUNet_NatureMethods_43} to standardize intensity ranges; 3) \textit{Segmentation:} Segmentation masks were generated using TotalSegmentator \cite{wasserthal2023totalsegmentator}, grouping 117 anatomical classes into 11 semantic categories (\eg, skeleton, muscle, cardioVascular; see Supplementary Material). Note that masks can also be manually annotated or automatically generated by other methods, ensuring flexibility for clinical use; 4) \textit{Transfer Functions:} We defined RGBA transfer functions for the 11 semantic groups to provide base appearance cues, enhancing the model's ability to predict Gaussian parameters; and 5) \textit{Ground-Truth Rendering:} For each training scan, we rendered 60 views of all classes, 30 views of skeleton group, and 30 views of organ groups (except the skeleton and muscle groups) with the resolution of 1600 $\times$ 1600 using physically-based rendering via PBRT \cite{pharr2023physically} to serve as ground truth, which takes 18.8 seconds per view on average. To enhance model robustness, we applied data augmentations, including random intensity shifts, Gaussian noise, and simulated acquisition artifacts, mimicking variations in clinical CT imaging.

For evaluation, we applied identical preprocessing to both test sets: 46 TotalSegmentator scans (\textit{ID}) and 10 CT-ORG scans (\textit{OOD}). Our experiments evaluated performance under three conditions: 1) \textit{Seen TF}: using transfer functions identical to training for testing generalization to new scans; 2) \textit{Unseen TF}: using novel transfer functions not seen during training to test appearance generalizability; and 3) \textit{Skeleton group}: visualizing only skeletal structures to evaluate compositional capabilities. For each test scan, we rendered 40 views for computing evaluation metrics, while 20 views were used for training 6DGS baseline or fine-tuning \modelname{}.

For semantic labeling of the 6DGS baseline, the original implementation does not support per-Gaussian class assignment. We therefore extended it using $k$-nearest neighbors (with $k$=1) to assign anatomical classes from the segmentation mask to each Gaussian primitive during initialization. In contrast, our anatomy-guided priming approach (6DGS + AGP) integrates semantic classification directly during the initialization process, providing more consistent anatomical structure representation.

\vspace{0.4em}
\noindent\textbf{Implementation Details} $\,$
\modelname{} was implemented in PyTorch using the Adam optimizer \cite{kingma2014adam} ($\beta_1=0.9$, $\beta_2=0.999$) with an initial learning rate of $1 \times 10^{-3}$ and PolyLR scheduling \cite{chen2017deeplab}. Due to GPU memory constraints, we used a batch size of 3 volumes, leveraging automatic mixed precision (AMP) and gradient accumulation for efficiency. Training was performed on a single NVIDIA A100 80GB GPU, requiring approximately 3 days. At inference, we additionally adopt FlashGS \cite{feng2024flashgs} to accelerate rasterization throughput. The number of instantiated Gaussians varied by scan complexity, ranging from 50,000 to 800,000, depending on the anatomical structures present.

The 6DGS experiments, including the original 6DGS and 6DGS with our AGP initialization (\ie, 6DGS + AGP), were trained following the official implementation \cite{6DGS_arxiv_39} for 30,000 iterations, with evaluations every 500 iterations to mitigate potential overfitting when training with limited views (20 per scan). We reported the best results for 6DGS experiments to ensure fair comparison. For \modelname{} fine-tuning (\textit{FT}), we conducted 300 optimization iterations, representing a substantial reduction in computational requirements, and reported the final results.

\begin{table}[tbp]
  \centering
  \caption{Quantitative comparison. \textit{ID}: in-domain (TotalSegmentator); \textit{OOD}: out-of-domain (CT-ORG); \textit{Seen/Unseen TF}: transfer functions used/not used for training; \textit{AGP}: anatomy-guided priming; \textit{Skeleton group}: compositional visualization requiring 0.0s additional preparation, enabled by per-Gaussian labels assigned during AGP.}
  \resizebox{\linewidth}{!}{
    \begin{tabular}{c|c|l|cccrrr}
    \toprule
    Dataset & Type & Method & SSIM  & PSNR  & LPIPS & \multicolumn{1}{c}{Time (s)} & \multicolumn{1}{c}{\# points} & \multicolumn{1}{c}{FPS} \\
    \midrule
   \multicolumn{1}{c|}{%
    \multirow{4}{*}{\rotatebox{90}{\texttt{TotalSeg}}}%
    }
    
    & \multicolumn{1}{c|}{\multirow{4}{*}{\makecell{ \textit{ID}\\\textit{Seen TF}}}} & 6DGS  & 0.912 & 26.63 & 0.096 & 1463.9 & 68,785 & 697.5 \\
          && 6DGS + AGP (\textbf{Ours}) & 0.925 & 28.92 & 0.093 & 1786.5 & 135,827 &  575.6 \\
          && \modelname{} (\textbf{Ours}) & 0.919 & 27.30 & 0.097 & \textbf{2.8} & 343,058 & 328.6 \\
          && \modelname{} (\textbf{Ours}) + FT & \textbf{0.937} & \textbf{31.67} & \textbf{0.088} & 89.4  & 227,925 &  423.8 \\
    \midrule
    \multicolumn{1}{c|}{\multirow{4}{*}{\begin{sideways} \texttt{CT-ORG} \end{sideways}}} & \multicolumn{1}{c|}{\multirow{4}{*}{\makecell{ \textit{OOD}\\\textit{Seen TF}}}} & 6DGS  & 0.903 & 25.97 & 0.105 & 1528.7 & 75,956 & 679.1 \\
          && 6DGS + AGP (\textbf{Ours}) & 0.926 & 29.36 & 0.091 & 2261.9 & 239,609 &   411.0\\
          && \modelname{} (\textbf{Ours}) & 0.918 & 26.21 & 0.092 & \textbf{2.6} & 586,225 & 245.2 \\
          && \modelname{} (\textbf{Ours}) + FT & \textbf{0.940} & \textbf{32.48} & \textbf{0.082} & 136.2 & 469,969 & 275.1 \\
    \midrule
    \multicolumn{1}{c|}{\multirow{4}{*}{\begin{sideways} \texttt{CT-ORG} \end{sideways}}} & \multicolumn{1}{c|}{\multirow{4}{*}{\makecell{ \textit{OOD}\\\textit{Unseen TF}}}} & 6DGS  & 0.908 & 26.59 & 0.103 & 1546.4 & 81,802 &  684.3\\
          && 6DGS + AGP (\textbf{Ours}) & 0.924 & 29.66 & 0.090 & 2129.3 & 246,017 &  385.5 \\
          && \modelname{} (\textbf{Ours}) & 0.913 & 26.77 & 0.092 & \textbf{2.8} & 586,225 &  245.5 \\
          && \modelname{} (\textbf{Ours}) + FT & \textbf{0.936} & \textbf{31.91} & \textbf{0.083} & 133.4 & 462,219 & 277.5 \\
    \midrule
    \multicolumn{1}{c|}{\multirow{4}{*}{\begin{sideways} \texttt{CT-ORG} \end{sideways}}} & \multicolumn{1}{c|}{\multirow{4}{*}{\makecell{ \textit{OOD}\\\textit{Seen TF}\\\textit{Skeleton group}}}} & 6DGS  & 0.935 & 26.78 & 0.066 & 5146.0 & 76,848 & 602.0 \\
          && 6DGS + AGP (\textbf{Ours}) & 0.938 & 28.95 & 0.064 & 7545.3 & 286,308 &   425.1 \\
          && \modelname{} (\textbf{Ours}) & 0.925 & 26.10 & 0.070 & \textbf{0.0} & 586,225 & 295.7  \\
          && \modelname{} (\textbf{Ours}) + FT & \textbf{0.944} & \textbf{30.74} & \textbf{0.061} & 140.1 & 466,638 & 337.3 \\

    \bottomrule
    \end{tabular}}%
  \label{tab:result}%
\end{table}%


\begin{figure*}[t]
    \centering
    \def\imgwidth{0.191\textwidth}
    \newcommand{\imgA}[1]{\includegraphics[width=\imgwidth,trim=92bp 122bp 92bp 61bp,clip]{#1}}  
    \newcommand{\imgB}[1]{\includegraphics[width=\imgwidth,trim=92bp 122bp 92bp 61bp,clip]{#1}}  
    \newcommand{\imgC}[1]{\includegraphics[width=\imgwidth,trim=77bp 153bp 77bp 0bp,clip]{#1}}  
    \newcommand{\imgD}[1]{\includegraphics[width=\imgwidth,trim=61bp 61bp 61bp 61bp,clip]{#1}}  
    \setlength{\tabcolsep}{0pt}
    \renewcommand{\arraystretch}{0.5}
    {\tiny
    \begin{tabular}{
    c
    @{\hskip 2pt} p{\imgwidth}
    @{\hskip 2pt} p{\imgwidth}
    @{\hskip 2pt} p{\imgwidth}
    @{\hskip 2pt} p{\imgwidth}
    @{\hskip 2pt} p{\imgwidth}}
    & \makebox[\imgwidth][c]{GT}
    & \makebox[\imgwidth][c]{6DGS}
    & \makebox[\imgwidth][c]{6DGS+AGP (\textbf{Ours})}
    & \makebox[\imgwidth][c]{Render-FM (\textbf{Ours})}
    & \makebox[\imgwidth][c]{Render-FM (\textbf{Ours})+FT} \\[2pt]
        \multirow{2}{*}[5em]{\rotatebox{90}{\textbf{TotalSeg} (\textit{ID, Seen TF})}}
        & \imgA{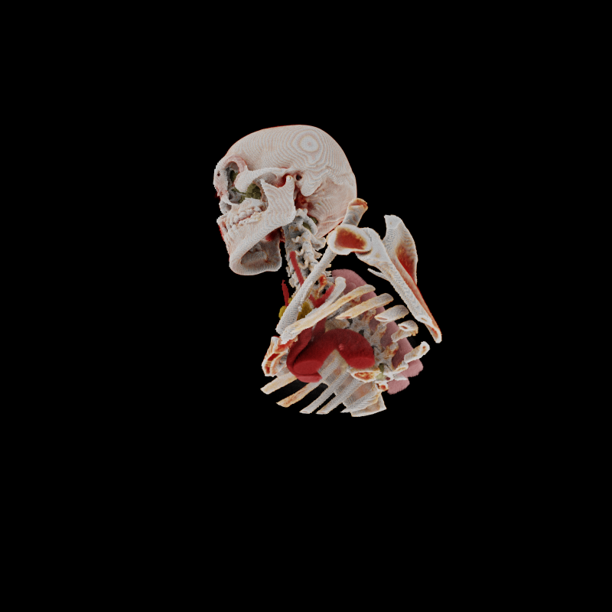}
        & \imgA{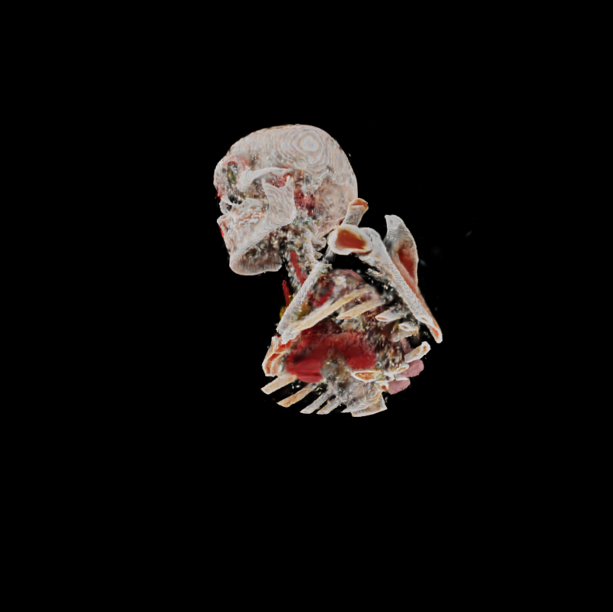}
        & \imgA{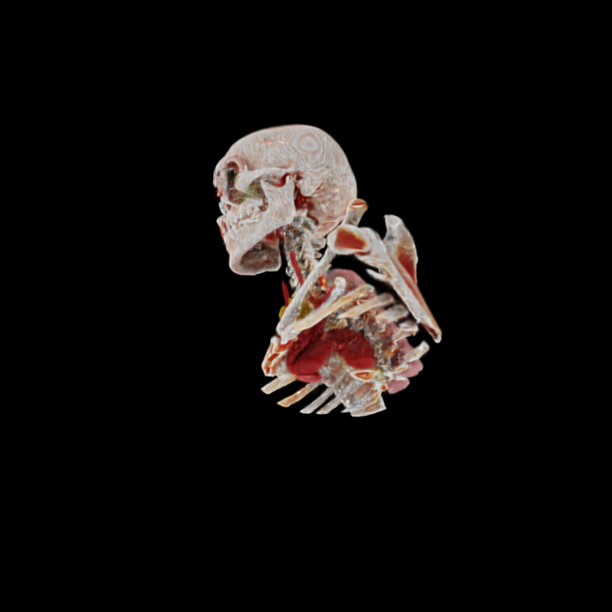}
        & \imgA{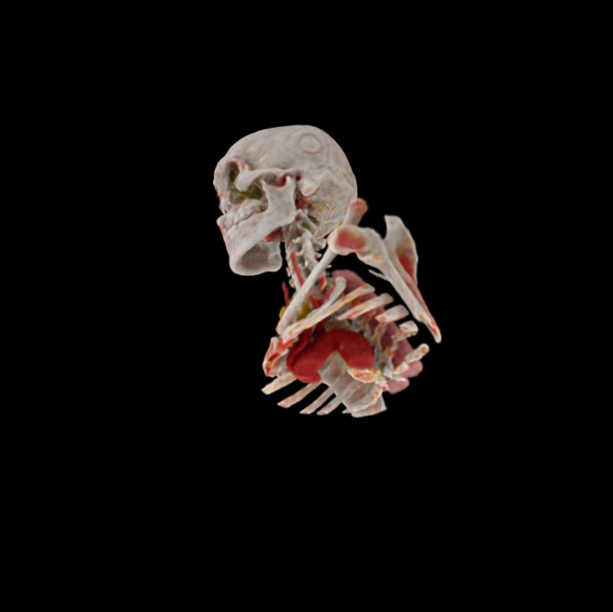}
        & \imgA{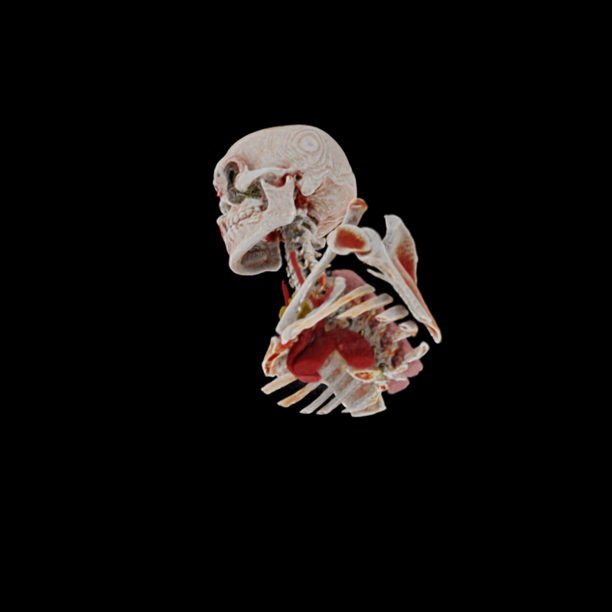} \\
        & \imgB{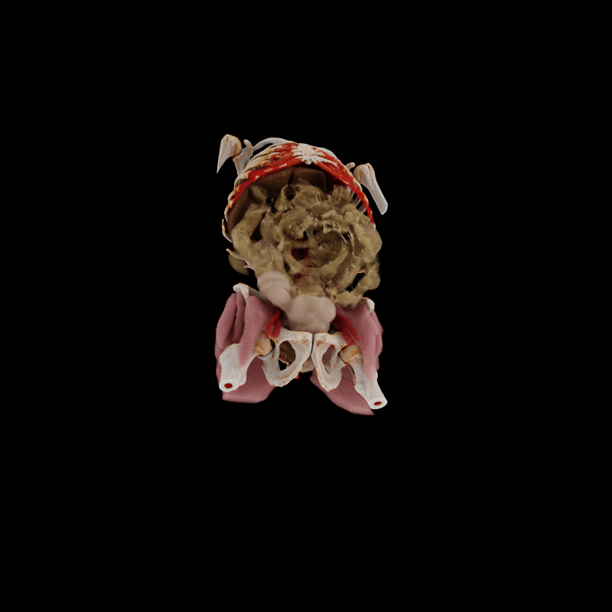}
        & \imgB{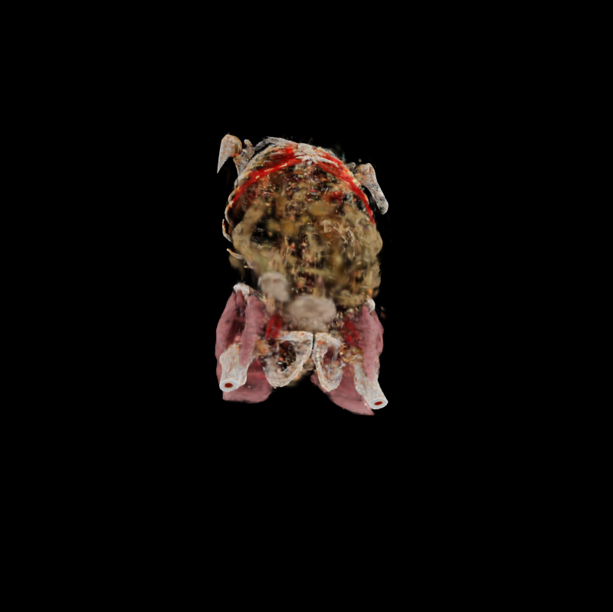}
        & \imgB{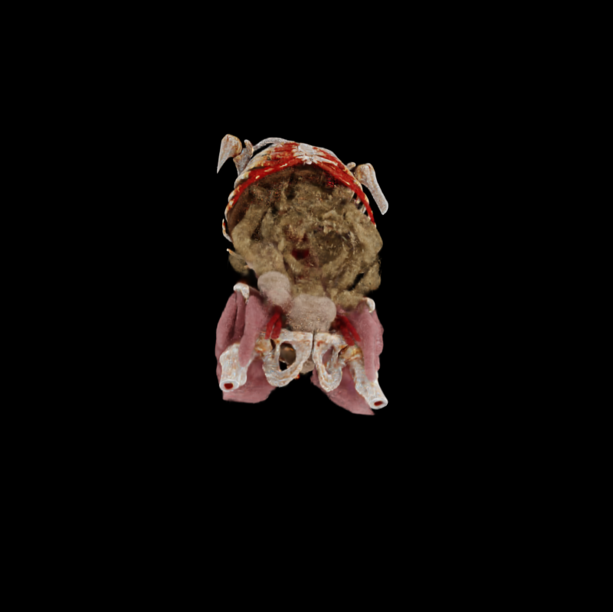}
        & \imgB{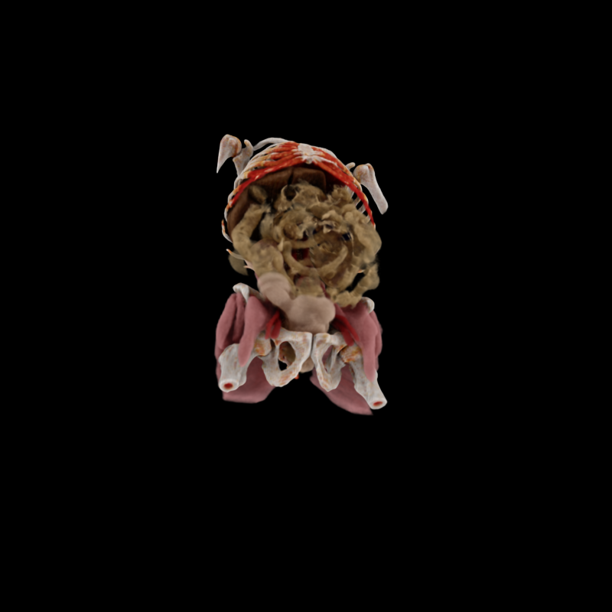}
        & \imgB{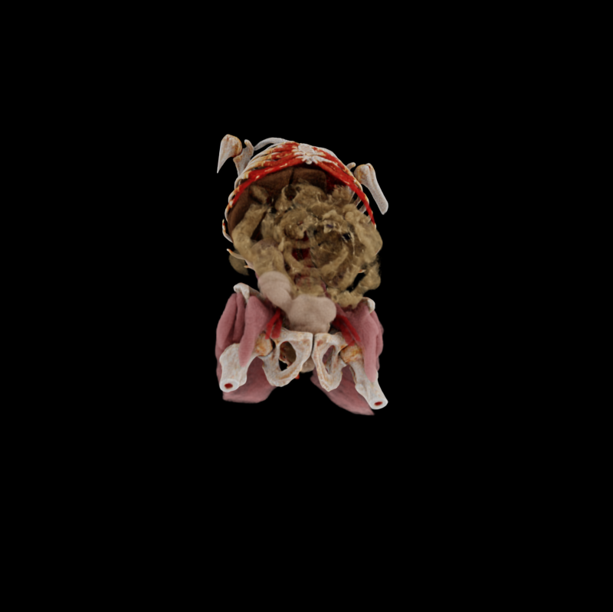} \\[2pt]
        \multicolumn{6}{c}{\tikz\draw[dashed, line width=0.5pt](0,0)--(\textwidth,0);} \\[2pt]
        \multirow{2}{*}[5em]{\rotatebox{90}{\textbf{CT-ORG} (\textit{OOD, Seen TF})}}
        & \imgC{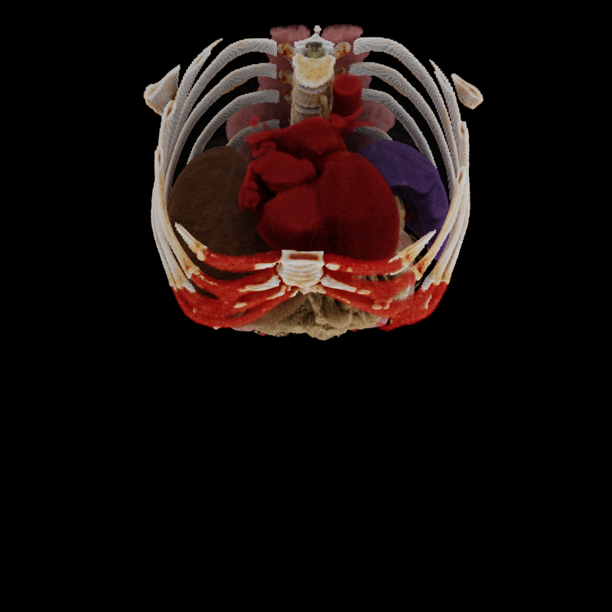}
        & \imgC{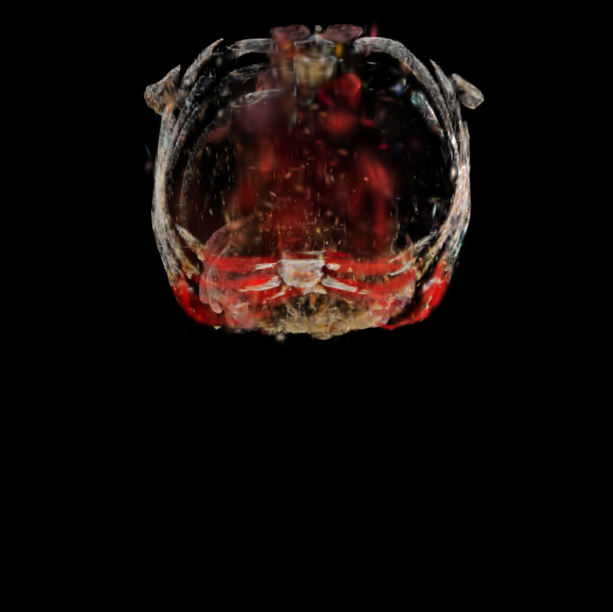}
        & \imgC{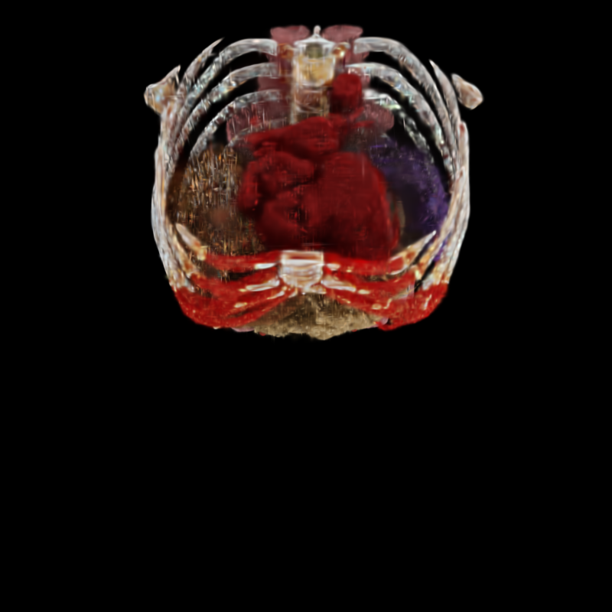}
        & \imgC{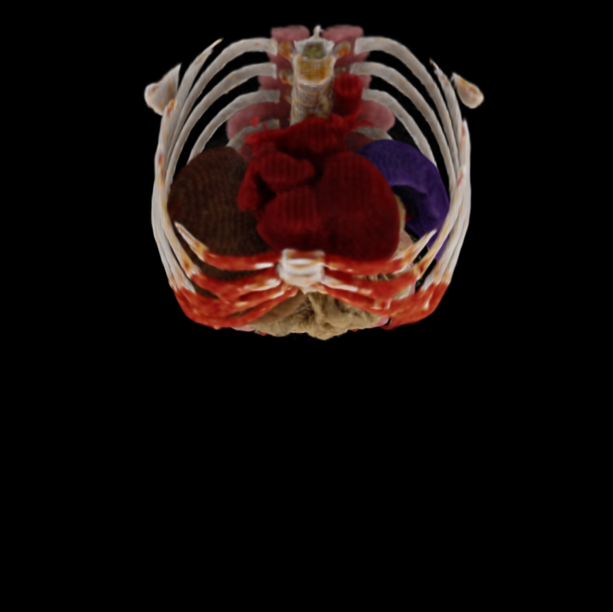}
        & \imgC{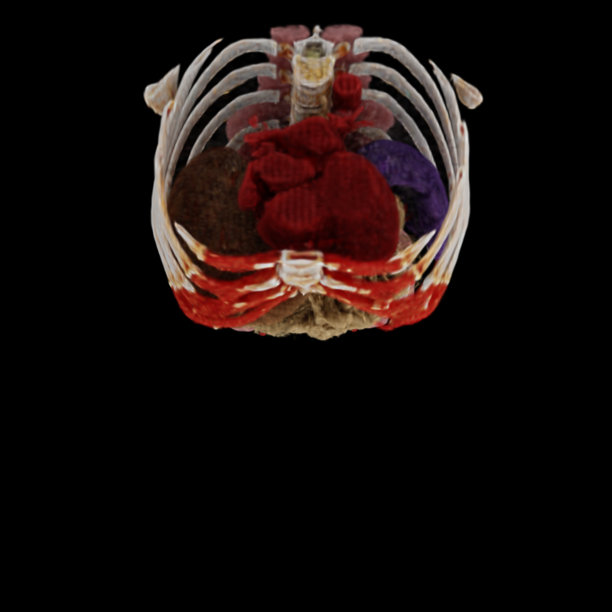} \\
        & \imgD{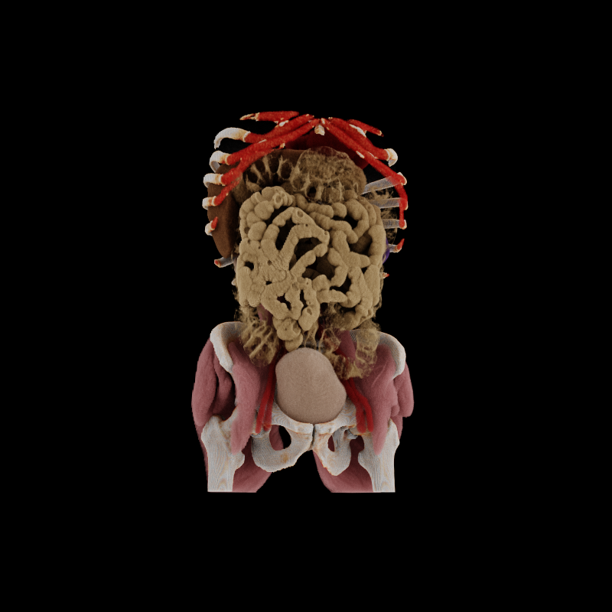}
        & \imgD{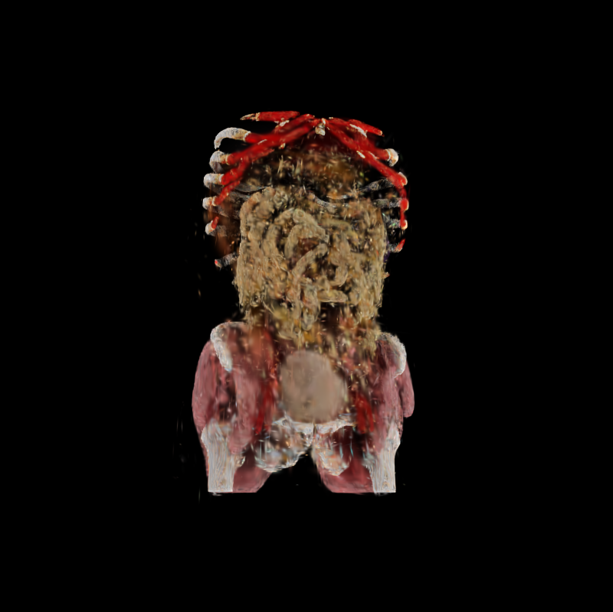}
        & \imgD{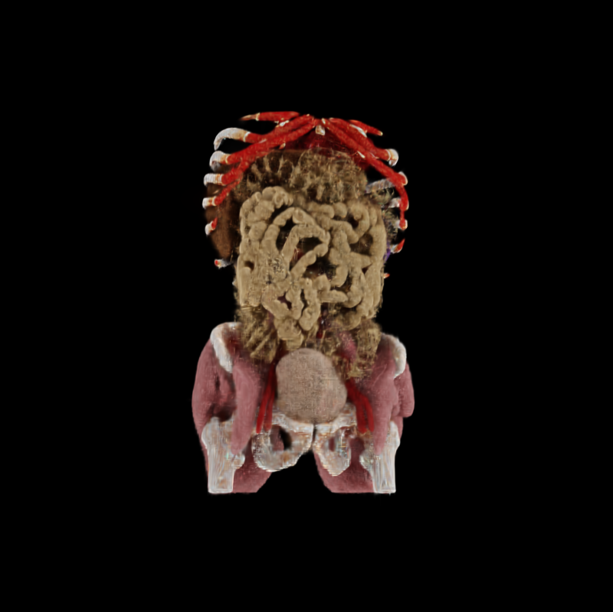}
        & \imgD{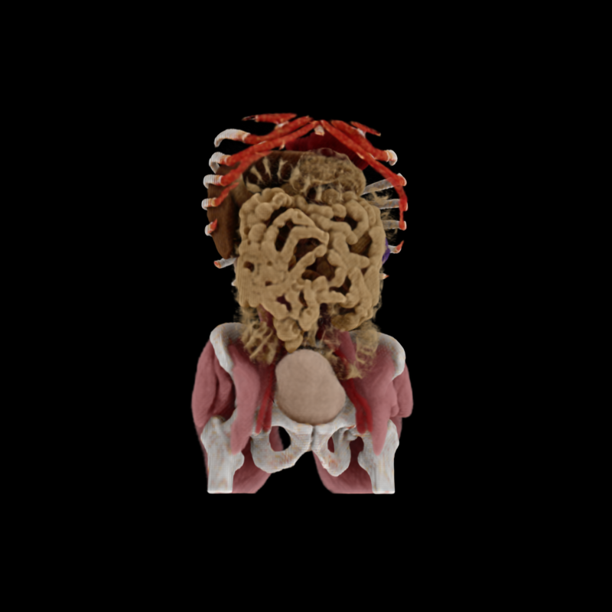}
        & \imgD{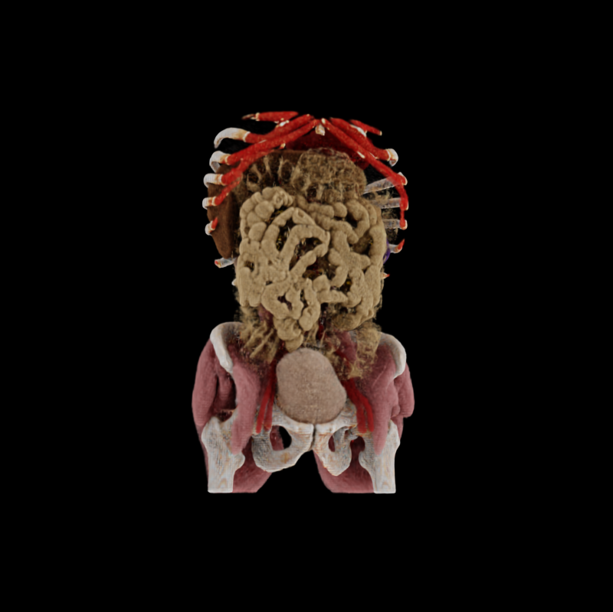} \\
    \end{tabular}
    }
    \caption{Qualitative comparison of methods on two datasets: in-domain (\textit{ID}) TotalSegmentator \cite{wasserthal2023totalsegmentator} and  out-of-domain (\textit{OOD}) CT-ORG \cite{rister2020ct} for seen transfer functions (\textit{Seen TF}).}
\label{fig:qualitative1}
\vspace{-2em}
\end{figure*}


\begin{figure*}[t]
    \centering
    \def\imgwidth{0.191\textwidth}
    \newcommand{\imgA}[1]{\includegraphics[width=\imgwidth,trim=92bp 183bp 92bp 0bp,clip]{#1}}  
    \newcommand{\imgB}[1]{\includegraphics[width=\imgwidth,trim=61bp 61bp 61bp 61bp,clip]{#1}}  
    \newcommand{\imgC}[1]{\includegraphics[width=\imgwidth,trim=61bp 61bp 61bp 61bp,clip]{#1}}  
    \newcommand{\imgD}[1]{\includegraphics[width=\imgwidth,trim=79bp 18bp 79bp 140bp,clip]{#1}}  
    \setlength{\tabcolsep}{0pt}
    \renewcommand{\arraystretch}{0.5}
    {\tiny
    \begin{tabular}{
    c
    @{\hskip 2pt} p{\imgwidth}
    @{\hskip 2pt} p{\imgwidth}
    @{\hskip 2pt} p{\imgwidth}
    @{\hskip 2pt} p{\imgwidth}
    @{\hskip 2pt} p{\imgwidth}}
    & \makebox[\imgwidth][c]{GT}
    & \makebox[\imgwidth][c]{6DGS}
    & \makebox[\imgwidth][c]{6DGS+AGP (\textbf{Ours})}
    & \makebox[\imgwidth][c]{Render-FM (\textbf{Ours})}
    & \makebox[\imgwidth][c]{Render-FM (\textbf{Ours})+FT} \\[2pt]
        \multirow{2}{*}[6em]{\rotatebox{90}{\textbf{CT-ORG} (\textit{OOD, Unseen TF})}}
        & \imgA{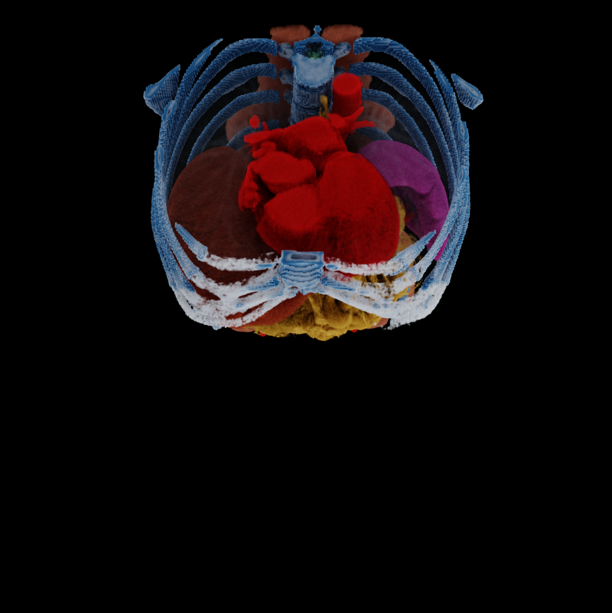}
        & \imgA{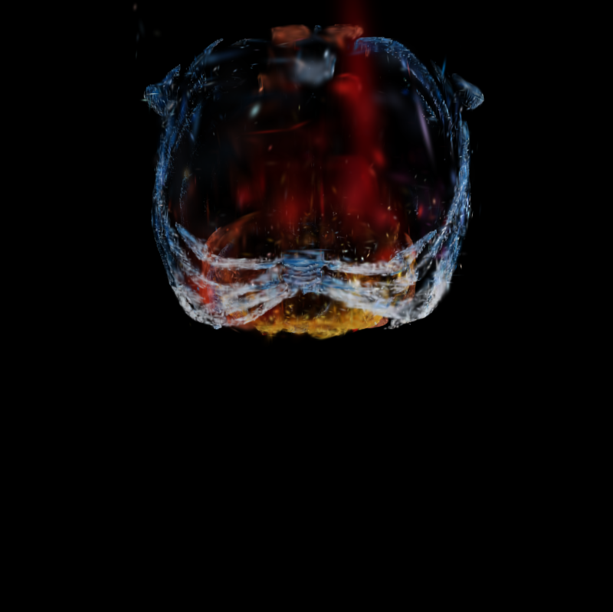}
        & \imgA{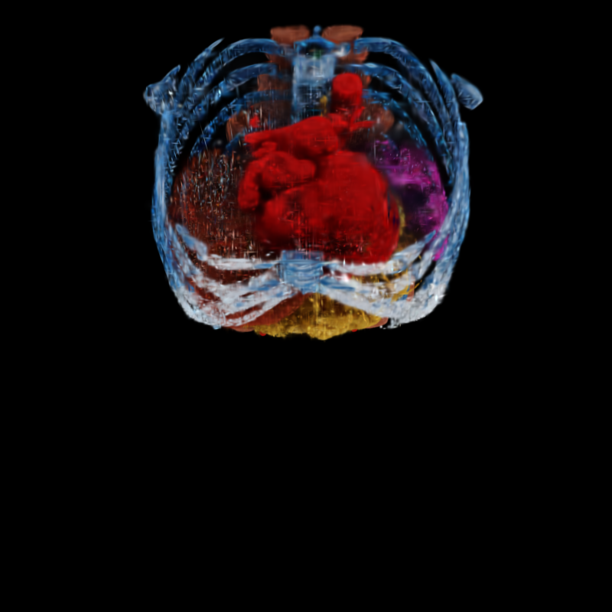}
        & \imgA{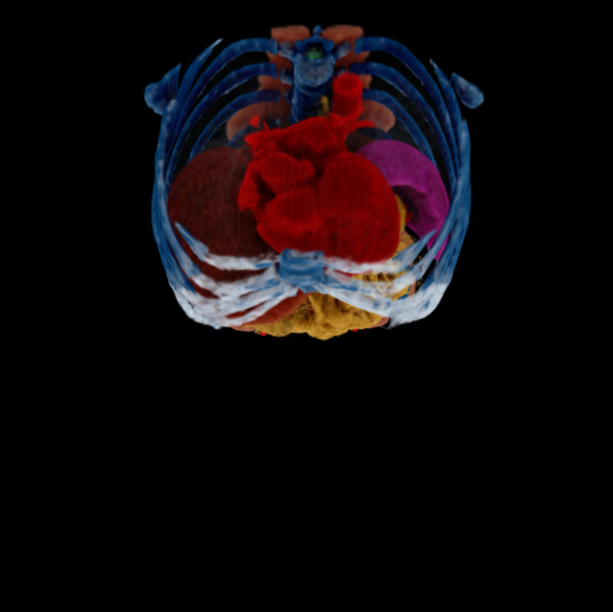}
        & \imgA{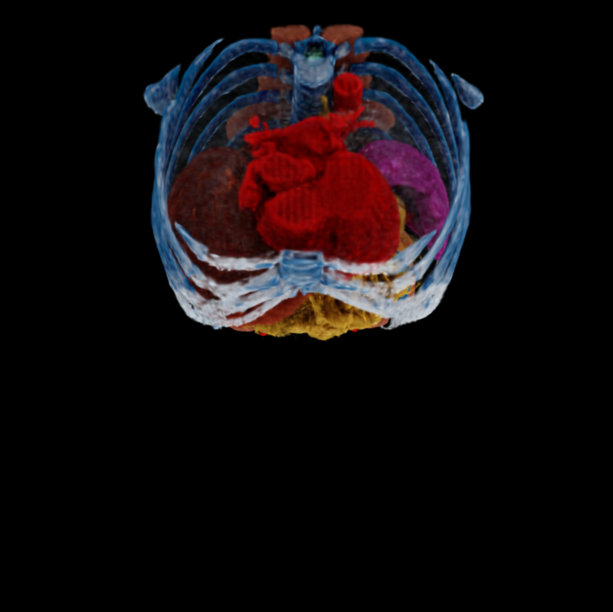} \\
        & \imgB{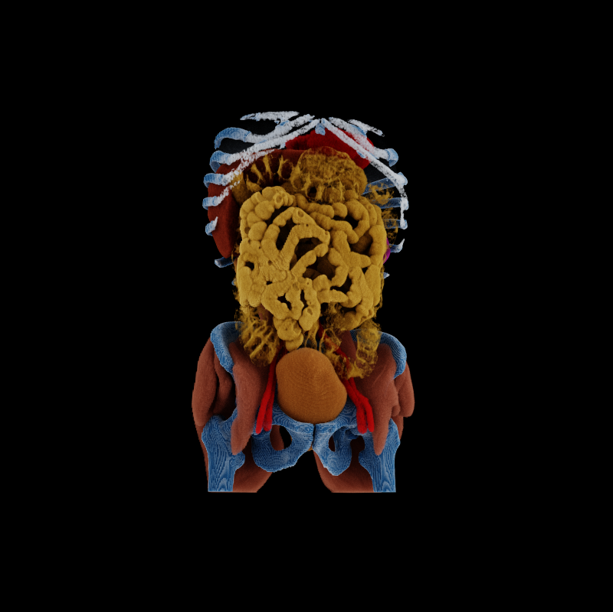}
        & \imgB{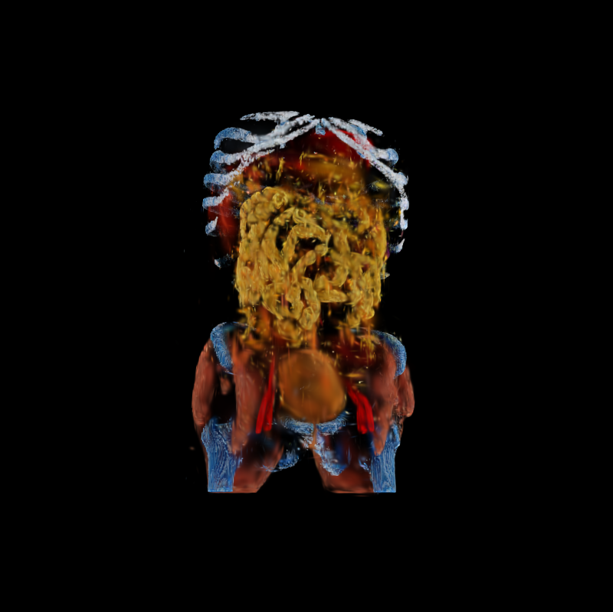}
        & \imgB{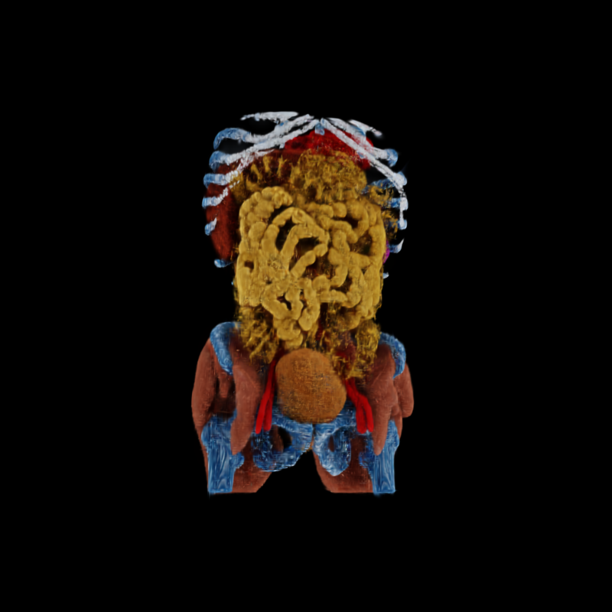}
        & \imgB{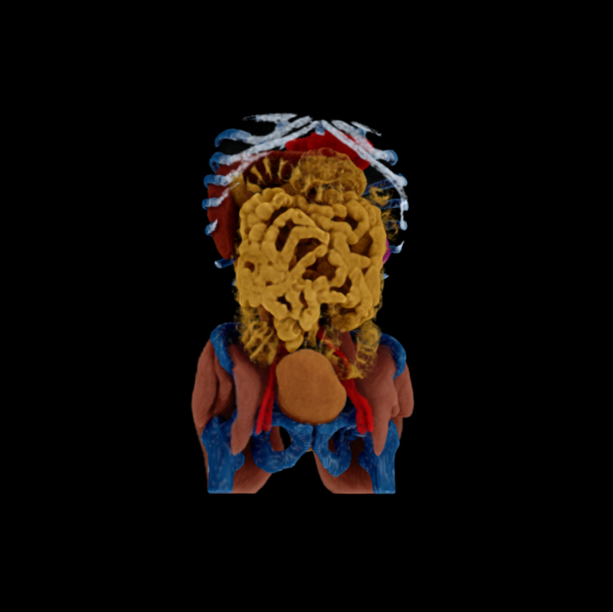}
        & \imgB{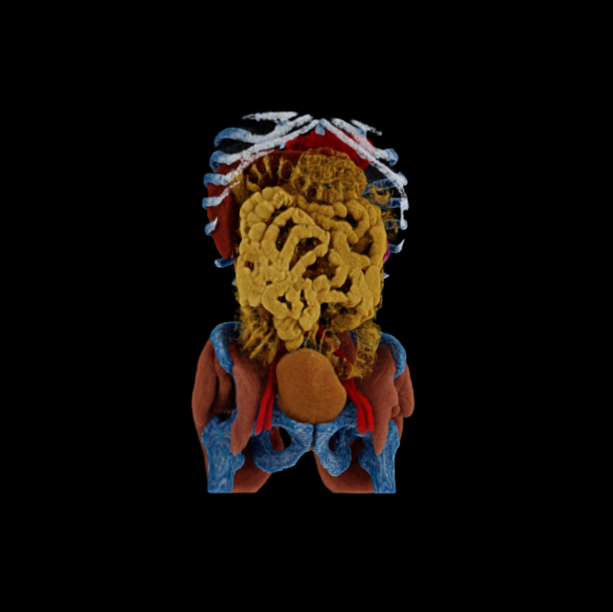} \\[2pt]
        \multicolumn{6}{c}{\tikz\draw[dashed, line width=0.5pt](0,0)--(\textwidth,0);} \\[2pt]
        \multirow{2}{*}[7em]{\rotatebox{90}{\textbf{CT-ORG} (\textit{OOD, Skeleton group})}}
        & \imgC{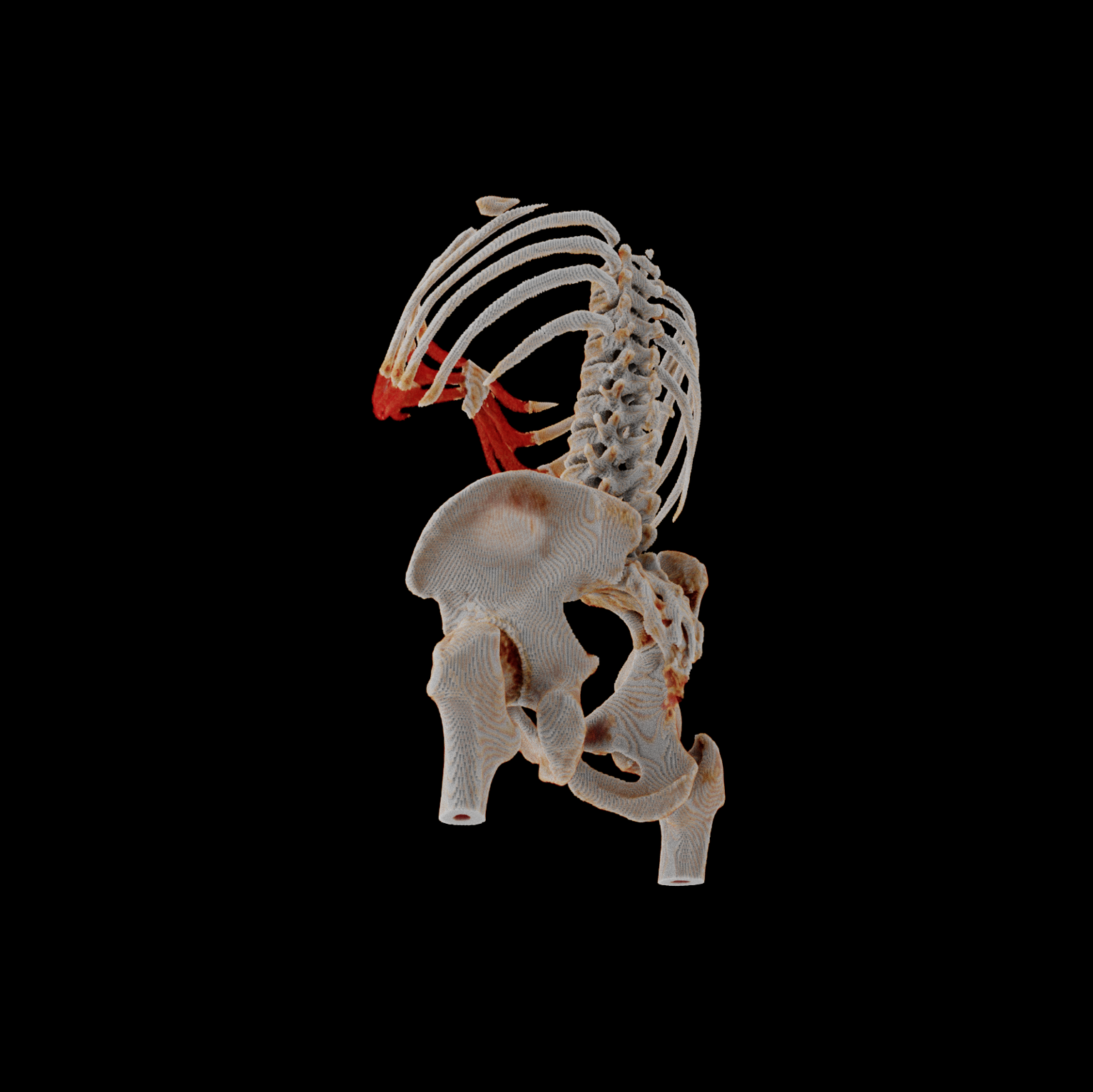}
        & \imgC{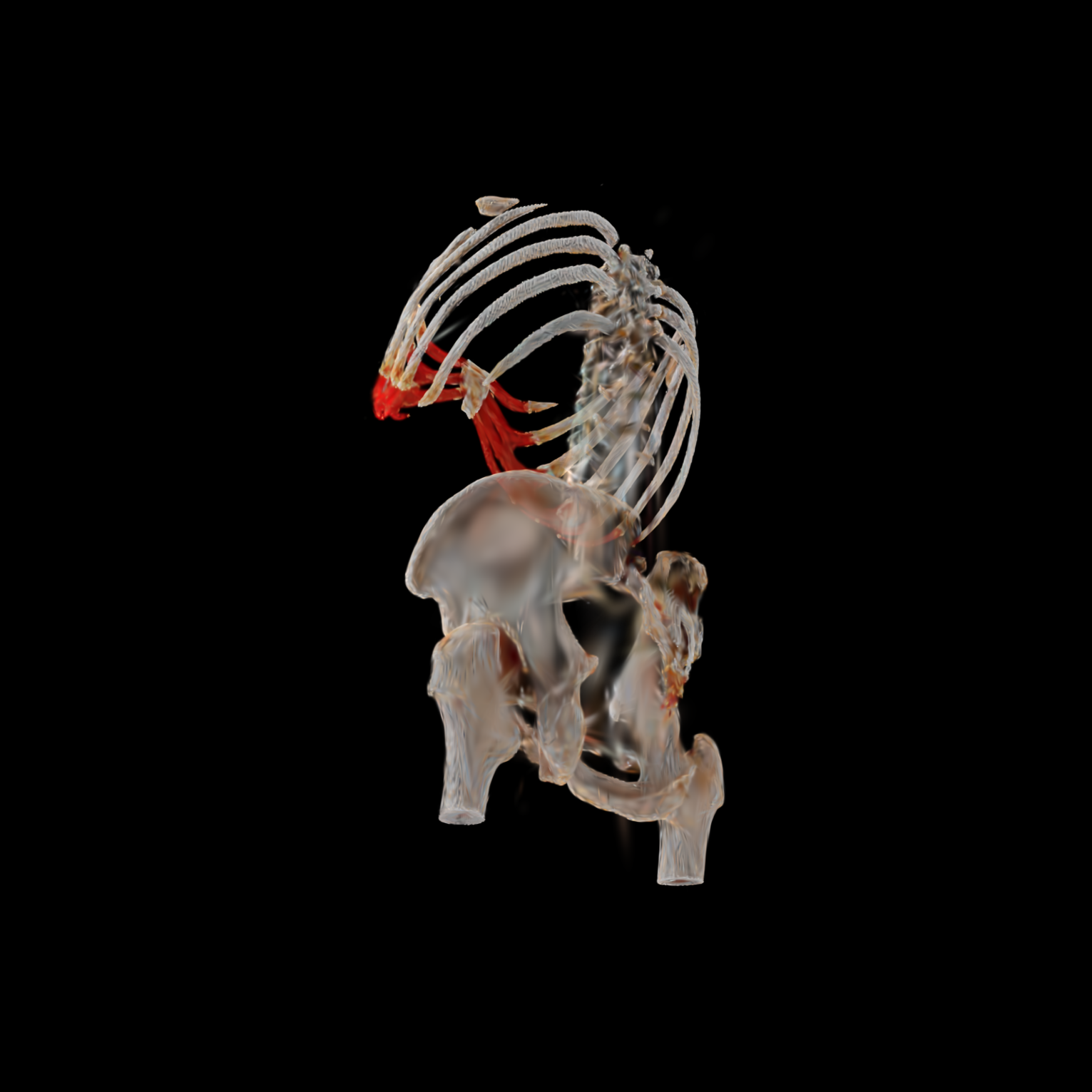}
        & \imgC{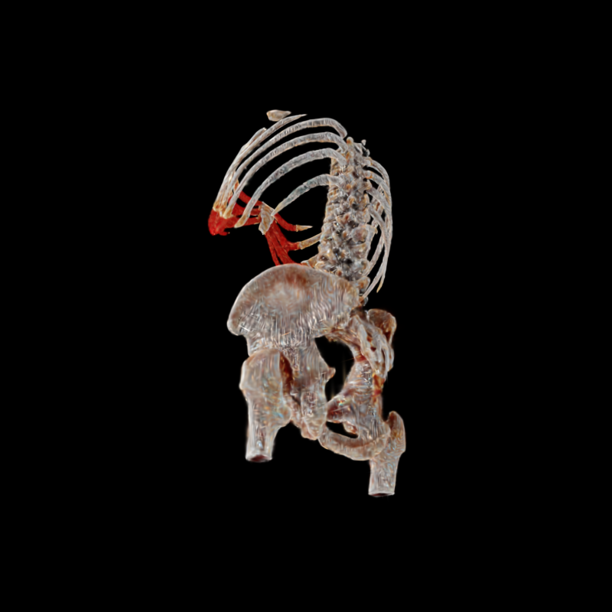}
        & \imgC{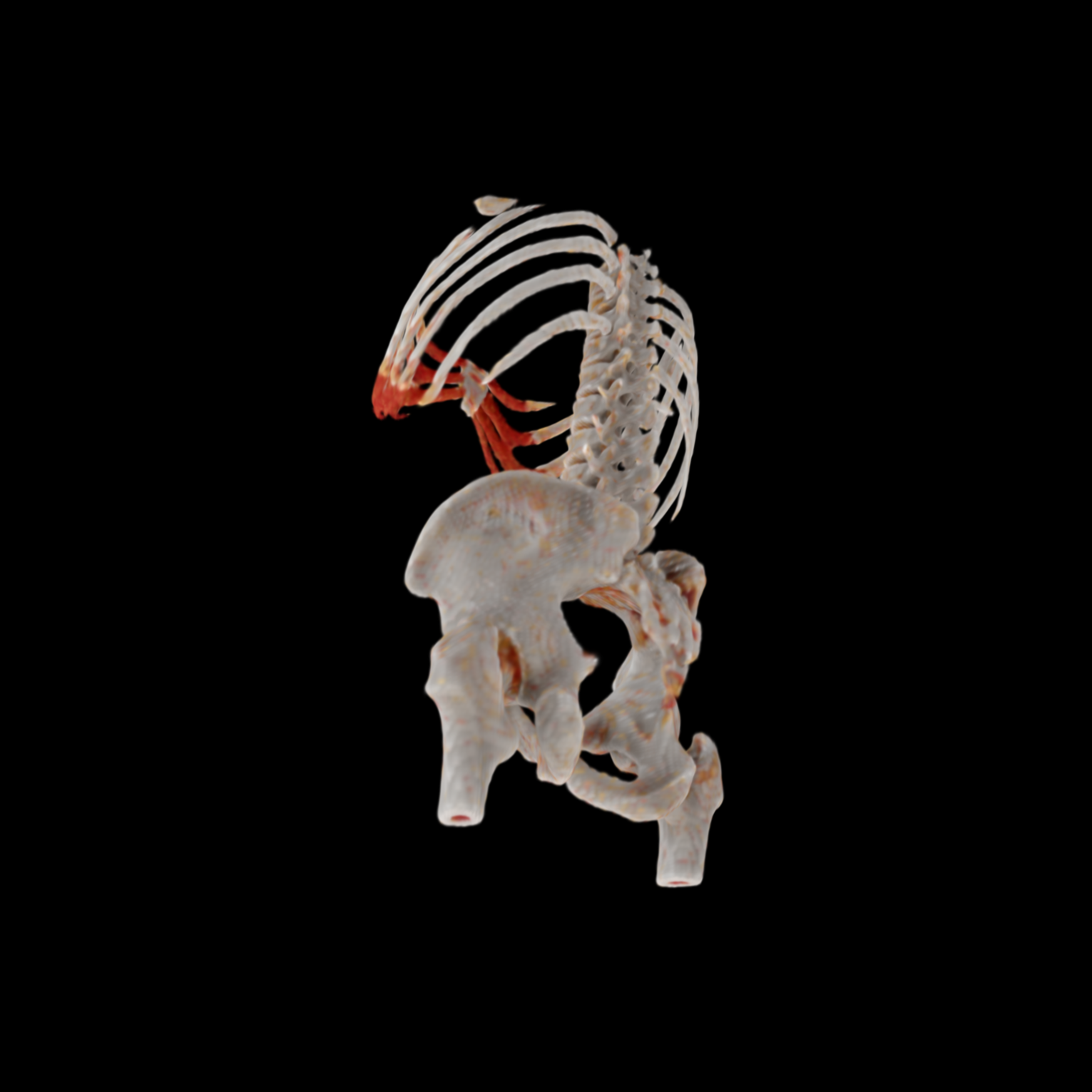}
        & \imgC{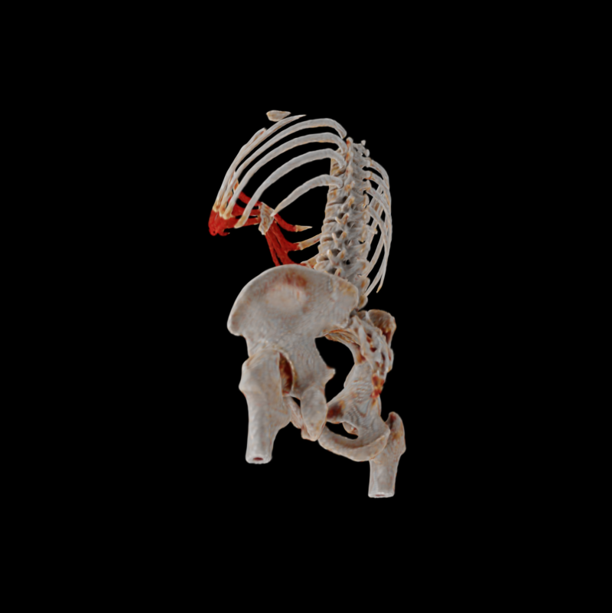} \\
        & \imgD{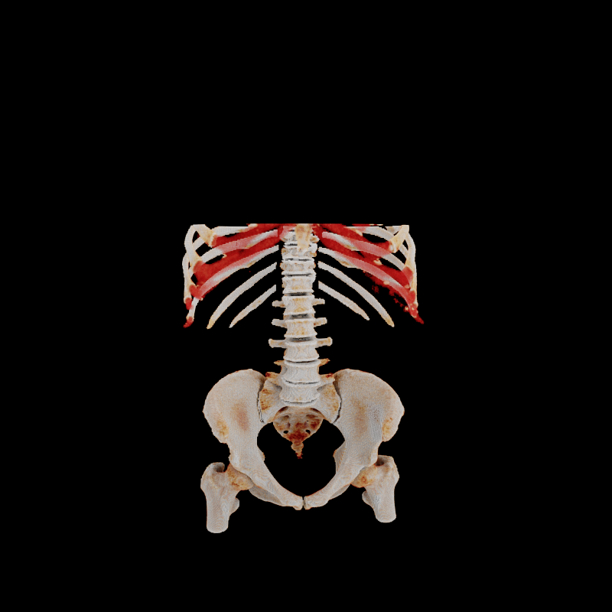}
        & \imgD{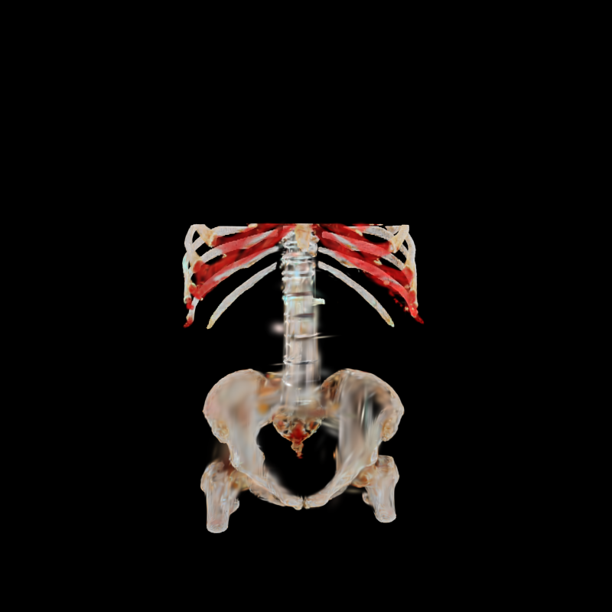}
        & \imgD{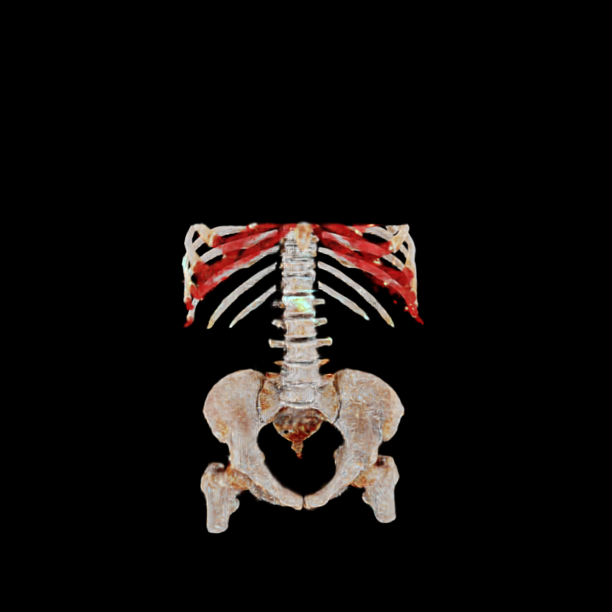}
        & \imgD{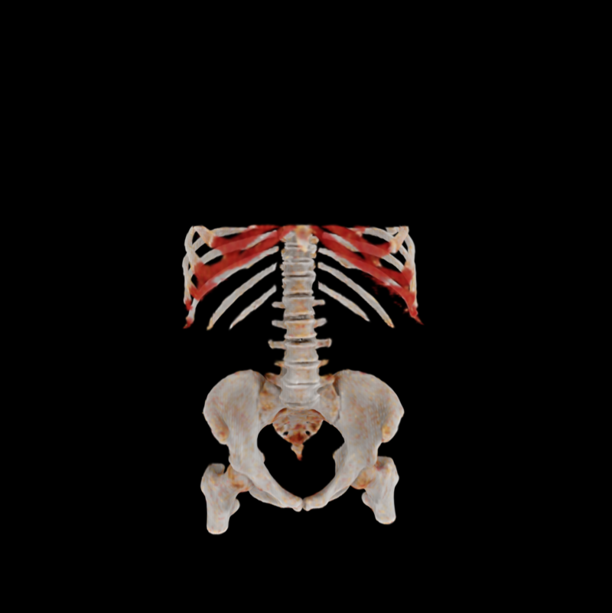}
        & \imgD{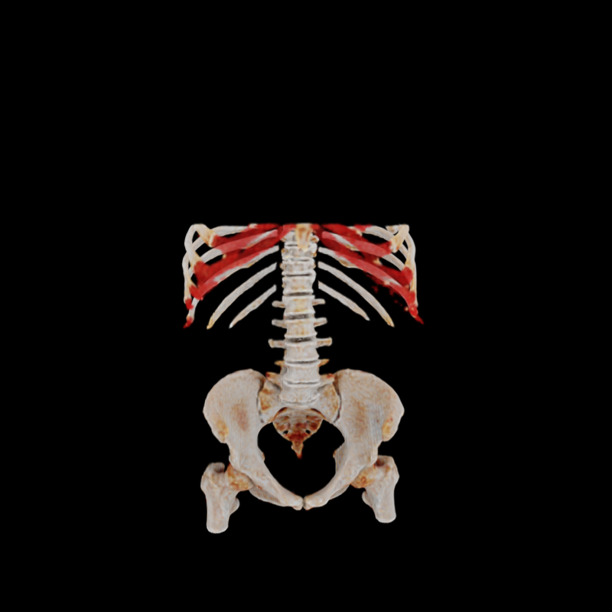} \\
    \end{tabular}
    }
    \caption{Qualitative comparison of methods on out-of-domain (\textit{OOD}) CT-ORG \cite{rister2020ct} for unseen transfer functions (\textit{Unseen TF}) and composability (\textit{Skeleton group}).}
\label{fig:qualitative2}
\vspace{-2em}
\end{figure*}

\vspace{0.4em}
\noindent\textbf{Evaluation Metrics} $\,$
To assess rendering quality, we employed three standard metrics: Peak Signal-to-Noise Ratio (PSNR), which quantifies pixel-level accuracy with higher values indicating better fidelity; Structural Similarity Index (SSIM) \cite{wang2004image}, which evaluates structural and perceptual similarity; and Learned Perceptual Image Patch Similarity (LPIPS) \cite{zhang2018unreasonable}, which measures perceptual similarity with lower values indicating closer alignment to human visual perception. For efficiency, we measured preparation time (in seconds), defined as the duration from raw CT input to 6DGS interactive rendering readiness; the number of Gaussian points, which reflects model complexity; and frames per second (FPS), which quantifies real-time rendering performance.

\subsection{Ablation: Anatomy-Guided Priming}
\label{sec:ablation}

We evaluate the contribution of Anatomy-Guided Priming (AGP) from two perspectives: its effect as an initialization strategy for per-scan optimization (6DGS + AGP vs. 6DGS), and its role as a necessary condition for feedforward learning in \modelname{}.

\vspace{0.4em}
\noindent\textbf{AGP as initialization for 6DGS} $\,$
As shown in Table \ref{tab:result}, applying AGP to the 6DGS baseline consistently improves rendering quality across all conditions. On TotalSegmentator (ID, Seen TF), 6DGS + AGP achieves SSIM of 0.925, PSNR of 28.92, and LPIPS of 0.093, compared to 0.912, 26.63, and 0.096 for vanilla 6DGS, with similar gains on CT-ORG under both seen (SSIM: 0.926 vs. 0.903) and unseen transfer functions (SSIM: 0.924 vs. 0.908). These improvements confirm that initializing Gaussians with anatomically-grounded positions, colors, and opacities from transfer functions provides a stronger starting point than marching-cubes. As a natural consequence of this richer initialization, fewer Gaussians are pruned during optimization, yielding a higher final point count (135,827 vs. 68,785 on TotalSegmentator) that better covers anatomical structures, which is a desired outcome reflecting improved scene representation. This also explains the longer optimization time (1786.5s vs. 1463.9s). Nonetheless, 6DGS + AGP still requires per-scan optimization for every new case.

\vspace{0.4em}
\noindent\textbf{AGP as a necessary component of \modelname{}} $\,$
For our Render-FM feedforward model, AGP is not merely beneficial but essential. Without AGP, \modelname{} fails to converge during training. This is because the feedforward network must predict meaningful 6DGS parameters for hundreds of thousands of voxels simultaneously, a highly underconstrained problem without a structured starting point. AGP resolves this by providing each Gaussian with a physically grounded initialization: positions anchored to voxel world coordinates, colors and opacities derived from transfer functions, and semantic labels from the segmentation mask. This transforms the learning objective from predicting absolute parameters from scratch into predicting residual corrections on top of meaningful priors, making the optimization landscape tractable. The convergence failure without AGP thus confirms that anatomy-guided initialization is a fundamental design requirement of \modelname{}, not an optional enhancement.

\subsection{Comparison with Baseline}
\label{sec:results}

Table \ref{tab:result} summarizes the performance of \modelname{}, the 6DGS baseline \cite{6DGS_arxiv_39}, and \modelname{} with fine-tuning (\textit{FT}) across the TotalSegmentator (\textit{ID}) \cite{wasserthal2023totalsegmentator} and CT-ORG (\textit{OOD}) \cite{rister2020ct} datasets, under \textit{Seen TF}, \textit{Unseen TF}, and \textit{Skeleton group} conditions.

\vspace{0.4em}
\noindent\textbf{In-Domain (\textit{TotalSeg, Seen TF})} $\,$
On the TotalSegmentator test set with seen transfer functions, \modelname{} surpasses the 6DGS baseline across all quality metrics (SSIM: 0.919 vs. 0.912, PSNR: 27.30 vs. 26.63, LPIPS: 0.097 vs. 0.096) without any per-scan optimization. Preparation time is reduced from 1463.9 seconds to just 2.8 seconds, a 500-fold improvement, while sustaining real-time rendering at 328.6 FPS. Fine-tuning further elevates performance substantially (SSIM: 0.937, PSNR: 31.67, LPIPS: 0.088), outperforming all methods at a preparation time of only 89.4 seconds.

\vspace{0.4em}
\noindent\textbf{Out-of-Domain (\textit{CT-ORG, Seen TF})} $\,$
For out-of-domain testing on CT-ORG with seen transfer functions, \modelname{} demonstrates robust generalization, outperforming 6DGS on all metrics (SSIM: 0.918 vs. 0.903, PSNR: 26.21 vs. 25.97, LPIPS: 0.092 vs. 0.105) with a preparation time of just 2.6 seconds at a rendering speed of 245.2 FPS. Fine-tuning significantly enhances performance further (SSIM: 0.940, PSNR: 32.48, LPIPS: 0.082), achieving the best results across all metrics with a preparation time of 136.2 seconds.

\vspace{0.4em}
\noindent\textbf{Unseen Transfer Functions (\textit{CT-ORG, Unseen TF})} $\,$
When evaluated with novel transfer functions unseen during training, \modelname{} maintains strong performance (SSIM: 0.913, PSNR: 26.77, LPIPS: 0.092), outperforming the 6DGS baseline (SSIM: 0.908, PSNR: 26.59, LPIPS: 0.103) and demonstrating that the learned representation generalizes to unseen appearance mappings without retraining. Fine-tuning further improves results (SSIM: 0.936, PSNR: 31.91, LPIPS: 0.083) with a preparation time of 133.4 seconds.

\vspace{0.4em}
\noindent\textbf{Composability (\textit{CT-ORG, Skeleton group})} $\,$
We further evaluated \modelname{}'s capability for compositional organ visualization. By selectively rendering Gaussians corresponding to specific anatomical classes (\eg, lungs, liver, bones) based on the segmentation mask, \modelname{} enables interactive exploration of individual or combined structures with zero additional preparation time (0.0s). This is a fundamental advantage over 6DGS: incorporating semantic labels into 6DGS requires a full separate per-scan optimization pass, inflating preparation time from 1528.7s to 5146.0s. \modelname{} achieves SSIM of 0.925, PSNR of 26.10, and LPIPS of 0.070 without any reprocessing, and fine-tuning further improves results to SSIM of 0.944, PSNR of 30.74, and LPIPS of 0.061 with only 140.1 seconds of preparation.

Across all conditions, \modelname{} consistently delivers rendering quality comparable to or better than per-scan optimized 6DGS, with preparation times reduced by two to three orders of magnitude. The dense Gaussian prediction strategy results in a higher point count and consequently lower FPS for \modelname{} (245.2--328.6 without fine-tuning; up to 423.8 with fine-tuning) compared to 6DGS (602.0--697.5), yet all values \textbf{far exceed} the threshold for real-time clinical interaction. Fine-tuning leverages \modelname{}’s robust initialization to achieve state-of-the-art quality with minimal additional computation.

Figures \ref{fig:qualitative1} and \ref{fig:qualitative2} provide a qualitative comparison across all methods. The 6DGS baseline, trained on 20 sparse views per scan, often overfits, producing floating noise artifacts in novel views. AGP-initialized 6DGS reduces these artifacts but does not fully eliminate them. In contrast, \modelname{} leverages its learned generalizable priors to largely mitigate floating noise, though it may appear slightly blurry in some cases. After brief fine-tuning, \modelname{} significantly enhances visual quality, capturing intricate details such as vasculature and bone interfaces with superior clarity, which are critical for clinical interpretation and diagnostic accuracy.

\section{Conclusion}
\label{sec:conclusion}

We presented \modelname{}, a feedforward model for real-time, high-fidelity volumetric rendering of CT scans using 6D Gaussian Splatting. By employing an nnU-Net inspired encoder-decoder architecture trained end-to-end, \modelname{} directly regresses 6DGS parameters from a multi-channel CT input volume, without the need for time-consuming per-scan optimization. The model leverages large-scale pre-training to learn generalizable mappings from CT data to expressive, view-dependent 3D representations. Our experiments demonstrate that \modelname{} achieves rendering quality comparable or better than that of per-scan optimized methods while reducing preparation time from hours to seconds, enabling interactive frame rates suitable for clinical use. By integrating the robustness of nnU-Net with the expressive power and view-dependent rendering capabilities of 6DGS, \modelname{} bridges the gap between advanced neural rendering quality and clinical practicality.

\vspace{0.4em}
\noindent\textbf{Future Work.} While this work focuses on CT, the \modelname{} pipeline is designed to be general: given modality-appropriate input channels and transfer functions, the same feedforward framework extends to other volumetric modalities such as MRI, which we leave as a promising direction for future work.


%
%
\bibliographystyle{splncs04}
\bibliography{main}

\clearpage
\setcounter{section}{0}
\setcounter{table}{0}
\setcounter{figure}{0}
\renewcommand{\thesection}{A\arabic{section}}
\renewcommand{\thetable}{A\arabic{table}}
\renewcommand{\thefigure}{A\arabic{figure}}

\begin{center}
{\Large\bfseries Supplementary Material}
\end{center}
\vspace{1em}


This appendix provides additional cross-dataset validation and implementation details for the \modelname{} framework, including comprehensive information on anatomical label consolidation from the TotalSegmentator dataset and detailed transfer function specifications for visualization. These details are intended to enhance reproducibility and provide deeper insights into our approach.

\section{Additional Cross-Dataset Validation: CTPelvic1K}

To further validate zero-shot generalization capability, we evaluate \modelname{} on the CTPelvic1K dataset \cite{ctpelvic1k}, an independent dataset from different institutions with distinct acquisition protocols and patient populations. Table \ref{tab:ctpelvic1k_results} presents quantitative results averaged over the first 6 scans, comparing our method against 6DGS baseline and 6DGS with AGP initialization. \modelname{} achieves competitive quality (SSIM: 0.926) in zero-shot inference with dramatic 2,500$\times$ speedup (0.65s vs. 1,643s for 6DGS). Note that the faster inference time (0.65s vs.\ 2.8s for TotalSegmentator) is due to the smaller average volume size of CTPelvic1K scans compared to the TotalSegmentator and CT-ORG datasets. With brief fine-tuning (78s), it achieves best quality across all metrics (SSIM: 0.941, PSNR: 33.30) while maintaining real-time rendering (553.6 FPS). These results confirm effective cross-dataset generalization without per-scan optimization.

\begin{table}[ht]
\centering
\caption{Quantitative results on CTPelvic1K dataset (first 6 scans average). \textit{AGP}: Anatomy-Guided Priming; \textit{FT}: Fine-tuning (300 iterations). All methods use the same 6DGS renderer backend.}
\label{tab:ctpelvic1k_results}
\begin{tabular}{lcccrrrc}
\toprule
\textbf{Method} & \textbf{SSIM} $\uparrow$ & \textbf{PSNR} $\uparrow$ & \textbf{LPIPS} $\downarrow$ & \textbf{Time (s)} $\downarrow$ & \textbf{\# Points} & \textbf{FPS} $\uparrow$ \\
\midrule
6DGS & 0.893 & 27.95 & 0.155 & 1,643.32 & 45,041 & 712.4 \\
6DGS + AGP (\textbf{Ours}) & 0.936 & 31.29 & 0.100 & 1,632.04 & 95,594 & 624.2 \\
\modelname{} (\textbf{Ours}) & 0.926 & 27.15 & 0.105 & \textbf{0.65} & 164,406 & 535.7 \\
\modelname{} (\textbf{Ours}) + FT & \textbf{0.941} & \textbf{33.30} & \textbf{0.096} & 78.02 & 149,243 & 553.6 \\
\bottomrule
\end{tabular}
\end{table}

\section{TotalSegmentator Label Mapping to Consolidated Groups}

Tables \ref{tab:label_mapping_part1} and \ref{tab:label_mapping_part2} detail the mapping of the original anatomical class labels from the TotalSegmentator dataset \cite{wasserthal2023totalsegmentator} (up to label 117) and user-defined labels (118, 119) to the 11 consolidated semantic groups used in the \modelname{} framework. This grouping strategy helps in managing the complexity of anatomical structures and provides a coherent basis for applying transfer functions and training the model.
The consolidated transfer function mappings are as follows:

\noindent\resizebox{\linewidth}{!}{%
\begin{minipage}{\linewidth}
\ttfamily\small
\vspace{1em}
// Consolidated Transfer Function Mappings:\\
// Index 0:\enspace Background/Other\\
// Index 1:\enspace Spleen\\
// Index 2:\enspace Liver\\
// Index 3:\enspace Digestive Group (Stomach, Bowels, Colon, GB, Panc, Eso)\\
// Index 4:\enspace Gland Group (Adrenals, Thyroid)\\
// Index 5:\enspace Lung Group\\
// Index 6:\enspace Trachea\\
// Index 7:\enspace Skeleton Group (Bones, Cartilage)\\
// Index 8:\enspace CardioVascular Group (Heart \& Vessels)\\
// Index 9:\enspace Nervous System Group (Brain, Spinal Cord)\\
// Index 10: Muscle Group\\
// Index 11: Kidney/Urogenital Group (Kidneys, Cysts, Bladder, Prostate)
\end{minipage}%
}

\section{Transfer Function Definitions}
Tables \ref{tab:tf_definitions_groups_0_5} and \ref{tab:tf_definitions_groups_6_11} detail the RGBA transfer functions used. Each transfer function is defined by a series of points, where each point has a Hounsfield Unit (HU) value and a corresponding RGBA (Red, Green, Blue, Alpha) value. The RGBA values are typically in the range of 0-255 for colors and 0.0-1.0 for alpha (opacity).

\begin{table}[th]
\centering
\caption{Mapping of TotalSegmentator and user-defined labels to the 11 consolidated semantic groups used in our framework (Part 1: labels 0--60).}
\label{tab:label_mapping_part1}
\resizebox{0.95\textwidth}{!}{%
\begin{tabular}{llcl}
\toprule
\textbf{Label ID} & \textbf{Class Name} & \textbf{Group ID} & \textbf{Group Name} \\
\midrule
0 & Background/Other & 0 & Background/Other \\ 
1 & spleen & 1 & Spleen \\ 
2 & kidney\_right & 11 & Kidney/Urogenital Group \\ 
3 & kidney\_left & 11 & Kidney/Urogenital Group \\ 
4 & gallbladder & 3 & Digestive Group \\ 
5 & liver & 2 & Liver \\ 
6 & stomach & 3 & Digestive Group \\ 
7 & pancreas & 3 & Digestive Group \\ 
8 & adrenal\_gland\_right & 4 & Gland Group \\ 
9 & adrenal\_gland\_left & 4 & Gland Group \\ 
10 & lung\_upper\_lobe\_left & 5 & Lung Group \\ 
11 & lung\_lower\_lobe\_left & 5 & Lung Group \\ 
12 & lung\_upper\_lobe\_right & 5 & Lung Group \\ 
13 & lung\_middle\_lobe\_right & 5 & Lung Group \\ 
14 & lung\_lower\_lobe\_right & 5 & Lung Group \\ 
15 & esophagus & 3 & Digestive Group \\ 
16 & trachea & 6 & Trachea \\ 
17 & thyroid\_gland & 4 & Gland Group \\ 
18 & small\_bowel & 3 & Digestive Group \\ 
19 & duodenum & 3 & Digestive Group \\ 
20 & colon & 3 & Digestive Group \\ 
21 & urinary\_bladder & 11 & Kidney/Urogenital Group \\ 
22 & prostate & 11 & Kidney/Urogenital Group \\ 
23 & kidney\_cyst\_left & 11 & Kidney/Urogenital Group \\ 
24 & kidney\_cyst\_right & 11 & Kidney/Urogenital Group \\ 
25 & sacrum & 7 & Skeleton Group \\ 
26 & vertebrae\_S1 & 7 & Skeleton Group \\ 
27 & vertebrae\_L5 & 7 & Skeleton Group \\ 
28 & vertebrae\_L4 & 7 & Skeleton Group \\ 
29 & vertebrae\_L3 & 7 & Skeleton Group \\ 
30 & vertebrae\_L2 & 7 & Skeleton Group \\ 
31 & vertebrae\_L1 & 7 & Skeleton Group \\ 
32 & vertebrae\_T12 & 7 & Skeleton Group \\ 
33 & vertebrae\_T11 & 7 & Skeleton Group \\ 
34 & vertebrae\_T10 & 7 & Skeleton Group \\ 
35 & vertebrae\_T9 & 7 & Skeleton Group \\ 
36 & vertebrae\_T8 & 7 & Skeleton Group \\ 
37 & vertebrae\_T7 & 7 & Skeleton Group \\ 
38 & vertebrae\_T6 & 7 & Skeleton Group \\ 
39 & vertebrae\_T5 & 7 & Skeleton Group \\ 
40 & vertebrae\_T4 & 7 & Skeleton Group \\ 
41 & vertebrae\_T3 & 7 & Skeleton Group \\ 
42 & vertebrae\_T2 & 7 & Skeleton Group \\ 
43 & vertebrae\_T1 & 7 & Skeleton Group \\ 
44 & vertebrae\_C7 & 7 & Skeleton Group \\ 
45 & vertebrae\_C6 & 7 & Skeleton Group \\ 
46 & vertebrae\_C5 & 7 & Skeleton Group \\ 
47 & vertebrae\_C4 & 7 & Skeleton Group \\ 
48 & vertebrae\_C3 & 7 & Skeleton Group \\ 
49 & vertebrae\_C2 & 7 & Skeleton Group \\ 
50 & vertebrae\_C1 & 7 & Skeleton Group \\ 
51 & heart & 8 & CardioVascular Group (Heart \& Vessels) \\ 
52 & aorta & 8 & CardioVascular Group (Heart \& Vessels) \\ 
53 & pulmonary\_vein & 8 & CardioVascular Group (Heart \& Vessels) \\ 
54 & brachiocephalic\_trunk & 8 & CardioVascular Group (Heart \& Vessels) \\ 
55 & subclavian\_artery\_right & 8 & CardioVascular Group (Heart \& Vessels) \\ 
56 & subclavian\_artery\_left & 8 & CardioVascular Group (Heart \& Vessels) \\ 
57 & common\_carotid\_artery\_right & 8 & CardioVascular Group (Heart \& Vessels) \\ 
58 & common\_carotid\_artery\_left & 8 & CardioVascular Group (Heart \& Vessels) \\ 
59 & brachiocephalic\_vein\_left & 8 & CardioVascular Group (Heart \& Vessels) \\ 
60 & brachiocephalic\_vein\_right & 8 & CardioVascular Group (Heart \& Vessels) \\ 
\bottomrule
\end{tabular}%
}
\end{table}

\begin{table}[th]
\centering
\caption{Mapping of TotalSegmentator and user-defined labels to the 11 consolidated semantic groups used in our framework (Part 2: labels 61--119).}
\label{tab:label_mapping_part2}
\resizebox{0.95\textwidth}{!}{%
\begin{tabular}{llcl}
\toprule
\textbf{Label ID} & \textbf{Class Name} & \textbf{Group ID} & \textbf{Group Name} \\
\midrule
61 & atrial\_appendage\_left & 8 & CardioVascular Group (Heart \& Vessels) \\ 
62 & superior\_vena\_cava & 8 & CardioVascular Group (Heart \& Vessels) \\ 
63 & inferior\_vena\_cava & 8 & CardioVascular Group (Heart \& Vessels) \\ 
64 & portal\_vein\_and\_splenic\_vein & 8 & CardioVascular Group (Heart \& Vessels) \\ 
65 & iliac\_artery\_left & 8 & CardioVascular Group (Heart \& Vessels) \\ 
66 & iliac\_artery\_right & 8 & CardioVascular Group (Heart \& Vessels) \\ 
67 & iliac\_vena\_left & 8 & CardioVascular Group (Heart \& Vessels) \\ 
68 & iliac\_vena\_right & 8 & CardioVascular Group (Heart \& Vessels) \\ 
69 & humerus\_left & 7 & Skeleton Group \\ 
70 & humerus\_right & 7 & Skeleton Group \\ 
71 & scapula\_left & 7 & Skeleton Group \\ 
72 & scapula\_right & 7 & Skeleton Group \\ 
73 & clavicula\_left & 7 & Skeleton Group \\ 
74 & clavicula\_right & 7 & Skeleton Group \\ 
75 & femur\_left & 7 & Skeleton Group \\ 
76 & femur\_right & 7 & Skeleton Group \\ 
77 & hip\_left & 7 & Skeleton Group \\ 
78 & hip\_right & 7 & Skeleton Group \\ 
79 & spinal\_cord & 9 & Nervous System Group (Brain, Spinal Cord) \\ 
80 & gluteus\_maximus\_left & 10 & Muscle Group \\ 
81 & gluteus\_maximus\_right & 10 & Muscle Group \\ 
82 & gluteus\_medius\_left & 10 & Muscle Group \\ 
83 & gluteus\_medius\_right & 10 & Muscle Group \\ 
84 & gluteus\_minimus\_left & 10 & Muscle Group \\ 
85 & gluteus\_minimus\_right & 10 & Muscle Group \\ 
86 & autochthon\_left & 10 & Muscle Group \\ 
87 & autochthon\_right & 10 & Muscle Group \\ 
88 & iliopsoas\_left & 10 & Muscle Group \\ 
89 & iliopsoas\_right & 10 & Muscle Group \\ 
90 & brain & 9 & Nervous System Group (Brain, Spinal Cord) \\ 
91 & skull & 7 & Skeleton Group \\ 
92 & rib\_left\_1 & 7 & Skeleton Group \\ 
93 & rib\_left\_2 & 7 & Skeleton Group \\ 
94 & rib\_left\_3 & 7 & Skeleton Group \\ 
95 & rib\_left\_4 & 7 & Skeleton Group \\ 
96 & rib\_left\_5 & 7 & Skeleton Group \\ 
97 & rib\_left\_6 & 7 & Skeleton Group \\ 
98 & rib\_left\_7 & 7 & Skeleton Group \\ 
99 & rib\_left\_8 & 7 & Skeleton Group \\ 
100 & rib\_left\_9 & 7 & Skeleton Group \\ 
101 & rib\_left\_10 & 7 & Skeleton Group \\ 
102 & rib\_left\_11 & 7 & Skeleton Group \\ 
103 & rib\_left\_12 & 7 & Skeleton Group \\ 
104 & rib\_right\_1 & 7 & Skeleton Group \\ 
105 & rib\_right\_2 & 7 & Skeleton Group \\ 
106 & rib\_right\_3 & 7 & Skeleton Group \\ 
107 & rib\_right\_4 & 7 & Skeleton Group \\ 
108 & rib\_right\_5 & 7 & Skeleton Group \\ 
109 & rib\_right\_6 & 7 & Skeleton Group \\ 
110 & rib\_right\_7 & 7 & Skeleton Group \\ 
111 & rib\_right\_8 & 7 & Skeleton Group \\ 
112 & rib\_right\_9 & 7 & Skeleton Group \\ 
113 & rib\_right\_10 & 7 & Skeleton Group \\ 
114 & rib\_right\_11 & 7 & Skeleton Group \\ 
115 & rib\_right\_12 & 7 & Skeleton Group \\ 
116 & sternum & 7 & Skeleton Group \\ 
117 & costal\_cartilages & 7 & Skeleton Group \\ 
118 & Coronary Arteries (User-defined) & 8 & CardioVascular Group (Heart \& Vessels) \\ 
119 & Pulmonary Artery (User-defined) & 8 & CardioVascular Group (Heart \& Vessels) \\ 
\bottomrule
\end{tabular}%
}
\end{table}

\begin{table}[th]
\centering
\caption{Definition of Transfer Functions (Groups 0--5): Seen TF and Unseen TF. For each consolidated group, the color theme is listed, followed by points defining HU values and their corresponding RGBA color and opacity.}
\label{tab:tf_definitions_groups_0_5}
\resizebox{1.0\textwidth}{!}{%
\begin{tabular}{@{}cl r l r l@{}}
\toprule
\textbf{Group ID} & \textbf{Group Name} & \multicolumn{2}{c}{\textbf{Seen TF}} & \multicolumn{2}{c}{\textbf{Unseen TF}} \\
\cmidrule(lr){3-4} \cmidrule(lr){5-6}
& & \textbf{Point (HU)} & \textbf{Value [R,G,B,A]} & \textbf{Point (HU)} & \textbf{Value [R,G,B,A]} \\
\midrule

\multirow{3}{*}{0} & \multirow{3}{*}{Background/Other}
 & \multicolumn{2}{c}{\textit{Neutral grayscale}} & \multicolumn{2}{c}{\textit{Neutral grayscale}} \\
 &  & -1024 & [0, 0, 0, 0] & -1024 & [0, 0, 0, 0] \\
 &  & 3072 & [0.0, 0.0, 0.0, 0.0] & 3072 & [0, 0, 0, 0] \\
\midrule

\multirow{8}{*}{1} & \multirow{8}{*}{Spleen}
 & \multicolumn{2}{c}{\textit{Soft purple gradient}} & \multicolumn{2}{c}{\textit{Vibrant Magenta/Purple}} \\
 &  & -1024 & [0, 0, 0, 0] & -1024 & [0, 0, 0, 0] \\
 &  & -150 & [0, 0, 0, 0] & 0 & [0, 0, 0, 0] \\
 &  & 20 & [70, 50, 90, 0.05] & 40 & [150, 40, 130, 0.1] \\
 &  & 80 & [110, 80, 140, 0.2] & 100 & [190, 70, 160, 0.3] \\
 &  & 180 & [150, 120, 170, 0.5] & 200 & [220, 100, 190, 0.6] \\
 &  & 250 & [190, 160, 200, 0.7] & 300 & [240, 130, 210, 0.8] \\
 &  & 3072 & [220, 190, 230, 0.85] & 3072 & [255, 160, 230, 0.85] \\
\midrule

\multirow{8}{*}{2} & \multirow{8}{*}{Liver}
 & \multicolumn{2}{c}{\textit{Realistic Brown Gradient}} & \multicolumn{2}{c}{\textit{Deep Red-Brown}} \\
 &  & -1024 & [0, 0, 0, 0] & -1024 & [0, 0, 0, 0] \\
 &  & -20 & [0, 0, 0, 0] & 10 & [0, 0, 0, 0] \\
 &  & 30 & [100, 70, 50, 0.1] & 50 & [130, 50, 30, 0.15] \\
 &  & 90 & [140, 100, 70, 0.3] & 120 & [160, 70, 50, 0.4] \\
 &  & 180 & [170, 130, 90, 0.6] & 220 & [180, 90, 70, 0.7] \\
 &  & 250 & [190, 150, 110, 0.75] & 300 & [195, 110, 85, 0.8] \\
 &  & 3072 & [210, 170, 130, 0.85] & 3072 & [210, 130, 100, 0.9] \\
\midrule

\multirow{8}{*}{3} & \multirow{8}{*}{Digestive Group}
 & \multicolumn{2}{c}{\textit{Beige/Brown (Realistic)}} & \multicolumn{2}{c}{\textit{Ochre/Yellow-Orange}} \\
 &  & -1024 & [0, 0, 0, 0] & -1024 & [0, 0, 0, 0] \\
 &  & -50 & [0, 0, 0, 0] & -20 & [0, 0, 0, 0] \\
 &  & 20 & [170, 140, 100, 0.05] & 30 & [190, 140, 50, 0.1] \\
 &  & 80 & [190, 160, 120, 0.25] & 90 & [210, 160, 70, 0.3] \\
 &  & 180 & [210, 180, 140, 0.55] & 190 & [230, 180, 90, 0.6] \\
 &  & 250 & [225, 195, 155, 0.7] & 280 & [245, 200, 110, 0.75] \\
 &  & 3072 & [240, 210, 170, 0.85] & 3072 & [255, 220, 130, 0.8] \\
\midrule

\multirow{8}{*}{4} & \multirow{8}{*}{Gland Group}
 & \multicolumn{2}{c}{\textit{Golden subtlety}} & \multicolumn{2}{c}{\textit{Muted Teal/Cyan}} \\
 &  & -1024 & [0, 0, 0, 0] & -1024 & [0, 0, 0, 0] \\
 &  & 0 & [0, 0, 0, 0] & 10 & [0, 0, 0, 0] \\
 &  & 30 & [160, 125, 35, 0.1] & 50 & [50, 120, 130, 0.15] \\
 &  & 100 & [200, 165, 70, 0.35] & 120 & [70, 150, 160, 0.4] \\
 &  & 200 & [220, 185, 80, 0.55] & 220 & [90, 180, 190, 0.65] \\
 &  & 250 & [240, 200, 90, 0.7] & 300 & [110, 200, 210, 0.75] \\
 &  & 3072 & [255, 225, 120, 0.75] & 3072 & [130, 220, 230, 0.8] \\
\midrule

\multirow{7}{*}{5} & \multirow{7}{*}{Lung Group}
 & \multicolumn{2}{c}{\textit{Realistic Pinkish Beige}} & \multicolumn{2}{c}{\textit{Very Light Airy Blue}} \\
 &  & -1024 & [0, 0, 0, 0] & -1024 & [0, 0, 0, 0] \\
 &  & -850 & [190, 180, 180, 0.0008] & -900 & [170, 190, 210, 0.001] \\
 &  & -500 & [210, 200, 200, 0.0025] & -600 & [190, 210, 230, 0.003] \\
 &  & 0 & [230, 220, 220, 0.004] & -100 & [210, 230, 245, 0.005] \\
 &  & 1000 & [240, 230, 230, 0.006] & 500 & [220, 240, 255, 0.007] \\
 &  & 3072 & [245, 235, 235, 0.008] & 3072 & [230, 245, 255, 0.009] \\
\bottomrule
\end{tabular}%
}
\end{table}

\begin{table}[th]
\centering
\caption{Definition of Transfer Functions (Groups 6--11): Seen TF and Unseen TF. For each consolidated group, the color theme is listed, followed by points defining HU values and their corresponding RGBA color and opacity.}
\label{tab:tf_definitions_groups_6_11}
\resizebox{1.0\textwidth}{!}{%
\begin{tabular}{@{}cl r l r l@{}}
\toprule
\textbf{Group ID} & \textbf{Group Name} & \multicolumn{2}{c}{\textbf{Seen TF}} & \multicolumn{2}{c}{\textbf{Unseen TF}} \\
\cmidrule(lr){3-4} \cmidrule(lr){5-6}
& & \textbf{Point (HU)} & \textbf{Value [R,G,B,A]} & \textbf{Point (HU)} & \textbf{Value [R,G,B,A]} \\
\midrule

\multirow{8}{*}{6} & \multirow{8}{*}{Trachea}
 & \multicolumn{2}{c}{\textit{Pale Beige/Pinkish}} & \multicolumn{2}{c}{\textit{Pale Lavender/Grey}} \\
 &  & -1024 & [0, 0, 0, 0] & -1024 & [0, 0, 0, 0] \\
 &  & -50 & [0, 0, 0, 0] & -80 & [0, 0, 0, 0] \\
 &  & 20 & [220, 200, 190, 0.1] & 0 & [180, 170, 190, 0.1] \\
 &  & 150 & [230, 210, 200, 0.35] & 100 & [200, 190, 210, 0.3] \\
 &  & 250 & [240, 220, 210, 0.5] & 200 & [220, 210, 230, 0.5] \\
 &  & 350 & [245, 225, 215, 0.65] & 350 & [235, 225, 245, 0.65] \\
 &  & 3072 & [250, 230, 220, 0.75] & 3072 & [245, 235, 255, 0.7] \\
\midrule

\multirow{7}{*}{7} & \multirow{7}{*}{Skeleton Group}
 & \multicolumn{2}{c}{\textit{Ivory bone realism}} & \multicolumn{2}{c}{\textit{Cool white to steel blue gradient}} \\
 &  & -1024 & [0, 0, 0, 0] & -1024 & [0.0, 0.0, 0.0, 0.0] \\
 &  & 100.0 & [180, 30, 30, 0.1] & 100.0 & [240.0, 248.0, 255.0, 0.0] \\
 &  & 180 & [255.0, 215.0, 140, 0.6] & 180 & [176.0, 196.0, 222.0, 0.8] \\
 &  & 280 & [255.0, 240.0, 240.0, 0.9] & 350 & [70.0, 130.0, 180.0, 1.0] \\
 &  & 350 & [255.0, 255.0, 255.0, 1.0] & 3072.0 & [70.0, 130.0, 180.0, 1.0] \\
 &  & 3072.0 & [255.0, 255.0, 255.0, 1.0] &  &  \\
\midrule

\multirow{9}{*}{8} & \multirow{9}{*}{CardioVascular Group}
 & \multicolumn{2}{c}{\textit{Muted Red Gradient}} & \multicolumn{2}{c}{\textit{Bright Anatomical Red}} \\
 &  & -1024 & [0, 0, 0, 0] & -1024 & [0, 0, 0, 0] \\
 &  & -50 & [0, 0, 0, 0] & 0 & [0, 0, 0, 0] \\
 &  & 50 & [120, 30, 30, 0.1] & 70 & [190, 20, 20, 0.2] \\
 &  & 150 & [160, 50, 50, 0.3] & 180 & [220, 40, 40, 0.5] \\
 &  & 250 & [180, 70, 70, 0.5] & 300 & [240, 60, 60, 0.75] \\
 &  & 400 & [200, 90, 90, 0.7] & 500 & [255, 80, 80, 0.85] \\
 &  & 600 & [220, 110, 110, 0.8] & 700 & [255, 120, 120, 0.9] \\
 &  & 3072 & [235, 150, 150, 0.85] & 3072 & [255, 150, 150, 0.95] \\
\midrule

\multirow{9}{*}{9} & \multirow{9}{*}{Nervous System Group}
 & \multicolumn{2}{c}{\textit{Soft beige/yellowish hues}} & \multicolumn{2}{c}{\textit{Soft Mint Green}} \\
 &  & -1024 & [0, 0, 0, 0] & -1024 & [0, 0, 0, 0] \\
 &  & -20 & [0, 0, 0, 0] & 0 & [0, 0, 0, 0] \\
 &  & 10 & [175, 165, 115, 0.1] & 30 & [120, 190, 140, 0.1] \\
 &  & 80 & [215, 205, 155, 0.35] & 100 & [150, 220, 170, 0.35] \\
 &  & 200 & [230, 220, 170, 0.5] & 220 & [180, 240, 200, 0.55] \\
 &  & 350 & [240, 230, 180, 0.7] & 400 & [200, 250, 220, 0.7] \\
 &  & 600 & [245, 235, 195, 0.75] & 700 & [220, 255, 235, 0.75] \\
 &  & 3072 & [255, 245, 225, 0.85] & 3072 & [235, 255, 245, 0.8] \\
\midrule

\multirow{8}{*}{10} & \multirow{8}{*}{Muscle Group}
 & \multicolumn{2}{c}{\textit{Realistic Muscle Pink to Beige}} & \multicolumn{2}{c}{\textit{Terracotta/Brownish-Red}} \\
 &  & -1024 & [0, 0, 0, 0] & -1024 & [0, 0, 0, 0] \\
 &  & 0 & [180, 120, 120, 0.05] & 20 & [160, 90, 70, 0.1] \\
 &  & 100 & [200, 140, 140, 0.25] & 120 & [180, 110, 90, 0.3] \\
 &  & 200 & [220, 160, 160, 0.4] & 220 & [200, 130, 110, 0.5] \\
 &  & 250 & [230, 170, 170, 0.55] & 350 & [215, 150, 130, 0.7] \\
 &  & 500 & [240, 180, 180, 0.7] & 600 & [230, 170, 150, 0.8] \\
 &  & 3072 & [245, 190, 190, 0.85] & 3072 & [240, 190, 170, 0.85] \\
\midrule

\multirow{8}{*}{11} & \multirow{8}{*}{Kidney/Urogenital Group}
 & \multicolumn{2}{c}{\textit{Beige Pink to Tan}} & \multicolumn{2}{c}{\textit{Warm Orange/Tan}} \\
 &  & -1024 & [0, 0, 0, 0] & -1024 & [0, 0, 0, 0] \\
 &  & 0 & [200, 170, 150, 0.05] & 15 & [190, 120, 60, 0.1] \\
 &  & 150 & [210, 180, 160, 0.35] & 100 & [210, 145, 80, 0.35] \\
 &  & 250 & [220, 190, 170, 0.55] & 200 & [225, 165, 100, 0.6] \\
 &  & 400 & [230, 200, 180, 0.7] & 350 & [240, 185, 120, 0.75] \\
 &  & 600 & [235, 205, 185, 0.75] & 600 & [250, 200, 140, 0.8] \\
 &  & 3072 & [240, 210, 190, 0.85] & 3072 & [255, 215, 160, 0.85] \\
\bottomrule
\end{tabular}%
}
\end{table}

\end{document}